\documentclass[runningheads]{llncs}

\usepackage[mobile]{eccv}

\usepackage{eccvabbrv}

\usepackage{graphicx}
\usepackage{booktabs}

\usepackage[accsupp]{axessibility}  

\usepackage[pagebackref,breaklinks,colorlinks,citecolor=eccvblue]{hyperref}

\usepackage{orcidlink}
\usepackage{xspace,mathtools,xcolor}
\usepackage{multirow}
\usepackage{wrapfig}
\usepackage{algorithm}
\usepackage{algpseudocode}

\newcommand{\vct}[1]{\boldsymbol{#1}\xspace}
\newcommand{\mat}[1]{\mathtt{#1}\xspace}

\begin{document}

\title{Robust Gaussian Splatting} 

\author{Fran\c{c}ois Darmon \quad Lorenzo Porzi \quad Samuel Rota-Bul\`{o} \quad Peter Kontschieder}

\authorrunning{Darmon \etal}
\institute{Meta Reality Labs Zurich \\ \email{fra.darmon@gmail.com}, \email{\{porzi,rotabulo,pkontschieder\}@meta.com}}
\titlerunning{Robust Gaussian Splatting}

\maketitle

\begin{figure}
    \centering
    \includegraphics[width=\linewidth]{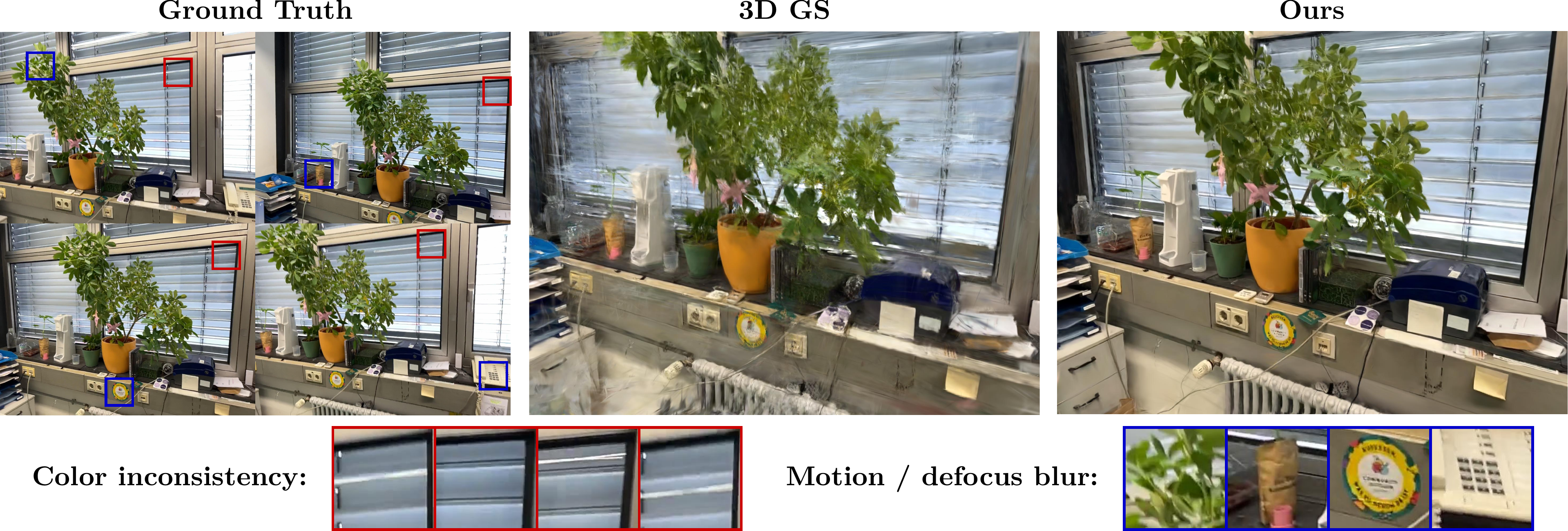}
    \caption{Hand-held phone captures (top left, from ScanNet++) can be challenging to reconstruct with 3D Gaussian Splatting (top center), due to inter-frame color inconsistencies (bottom left), motion blur and defocus blur (bottom right). We show how these factors can explicitly and easily be modeled in the 3D GS framework, leading to notably improved reconstruction results (top right).}
    \label{fig:teaser}
\end{figure}

\begin{abstract}
In this paper, we address common error sources for 3D Gaussian Splatting (3DGS) including blur, imperfect camera poses, and color inconsistencies, with the goal of improving its robustness for practical applications like reconstructions from handheld phone captures. Our main contribution involves modeling motion blur as a Gaussian distribution over camera poses, allowing us to address both camera pose refinement and motion blur correction in a unified way. Additionally, we propose mechanisms for defocus blur compensation and for addressing color inconsistencies caused by ambient light, shadows, or due to camera-related factors like varying white balancing settings. Our proposed solutions integrate in a seamless way with the 3DGS formulation while maintaining its benefits in terms of training efficiency and rendering speed. We experimentally validate our contributions on relevant benchmark datasets including Scannet++ and Deblur-NeRF, obtaining state-of-the-art results and thus consistent improvements over relevant baselines.

\end{abstract}

\newpage

\section{Introduction}
\label{sec:intro}




Neural rendering has emerged as a powerful tool for photorealistic novel view synthesis (NVS), the task of rendering images from arbitrary camera viewpoints, given a set of posed input images. NVS has a broad number of applications in fields such as AR/VR/MR, robotics, mapping, \etc. One of the most recently proposed methods are Neural Radiance Fields (NeRFs)~\cite{nerf}, which combine deep learning and volumetric rendering techniques for modelling 3D scenes. This is achieved by optimizing multilayer perceptrons (MLPs) to map from spatial coordinates and viewing directions to density and color fields, allowing the models to capture the intricate relationship between light and geometry in a data-driven manner, with remarkable fidelity.

Another recent approach to neural rendering is 3D Gaussian splats (3DGS)~\cite{kerbl3Dgaussians}, which represent scenes as a collection of anisotropic 3D Gaussian functions with corresponding, additional parameters for modelling view-dependent appearance and opacity. This representation allows for very fast and GPU-efficient rendering of high-quality, high-resolution images, making it well-suited for applications with strong fidelity requirements.

While the aforementioned neural rendering techniques have shown great promise in generating photorealistic images and videos, they often make strong assumptions about their input. For example, many methods assume perfect camera poses, non-blurred input images, and without any defocus artifacts. Additionally, they may require constant camera color settings and calibration, which can be difficult to achieve in practice. Such assumptions limit the applicability of neural rendering techniques for real-world scenarios, where input data may be noisy, incomplete, and distorted. As a consequence, images rendered from models trained on real-world data exhibit characteristic issues including floaters and poor image quality.

In this paper, we explore ways to relax the aforementioned data quality assumptions for 3D Gaussian splats and improve their robustness for practical applications. More specifically, we introduce a new formulation for addressing common error sources including blur (camera motion blur and defocus blur), imperfect camera poses, and color inconsistencies. 
We explain that motion blur in an image can be understood as the result of integrating multiple, locally perturbed camera captures taken during exposure time. 
Building on this observation, we propose a solution that addresses both camera pose distortion and motion blur by modeling motion blur as a Gaussian distribution over camera poses, from where we can obtain the expected image at the given noise. This approach seamlessly and elegantly integrates with the probabilistic formulation of 3DGS while maintaining its benefits in terms of training efficiency and rendering speed. For our model, we only require few additional, per-image parameters (6 for motion blur), based on which we can derive the updated per-Gaussian parameters. 

Defocus blur is caused by the use of an aperture in a camera, which limits the amount of light entering the lens. When the aperture is small, only a narrow beam of light can pass through the lens, resulting in a sharp image. However, when the aperture is larger, more light can pass through the lens, causing the image to become blurry, particularly under low-light conditions. Additionally, the use of an aperture also causes the depth of field to be shallow, meaning that only objects within a certain distance from the camera will be in focus (on the so-called focus plane). Conversely, objects outside of this range will appear blurry, and with increasing distance to the focus plane, get projected to a circle with increasing radius. To compensate for defocus blur, we propose an offset correction mechanism implemented via another covariance to the Gaussians, once projected to the 2D image plane.

To address color inconsistencies across multiple images taken for a scene that are caused by changes in ambient light, moving shadows, or camera-induced issues like white balancing, we propose incorporating an RGB decoder function with additional per-image parameters. This can be implemented as a simple affine color transformation applied to the standard RGB color decoder of 3D Gaussian Splats. For novel view synthesis at test time, this can be absorbed into the spherical harmonics used in conventionally used 3DGS viewers.

We experimentally validate our contributions and their impact on relevant benchmark datasets including the novel, real-world Scannet++~\cite{scannetpp} dataset and the synthetically generated Deblur-NeRF dataset~\cite{deblurnerf}. For Scannet++, we introduce an evaluation procedure for novel view synthesis, leveraging their noisy, low-quality iPhone data for evaluation. To address blur and other issues in the images, we select the 10 best test views per sequence based on a blurriness measure (decreasing maximum gradient magnitude), while ensuring no test view is within $0.5$m and $60^\circ$ of another. To address pose and color drift, we propose a test-time adaptation to align the camera poses and color profile to the expected one from ground truth before computing evaluation metrics. This approach does not lead to over-fitting and can be applied to all baselines in a fair way. Our experiments show consistent improvements over relevant baselines including 3DGS~\cite{kerbl3Dgaussians} and MipNeRF-360~\cite{mipnerf360}.

To summarize, our proposed contributions are as follows:
\begin{itemize}
    \item A solution that models motion blur as a Gaussian distribution over camera poses, allowing us to address both camera pose refinement and motion blur correction while maintaining the benefits of 3DGS in terms of training efficiency and rendering speed.
    \item We propose a defocus blur compensation mechanism, providing an off-focus plane correction implemented via another covariance to the Gaussians, once projected to the 2D image plane.
    \item We propose an RGB decoder function with per-image parameters to address color inconsistencies caused by ambient light, shadows, or camera-specific issues.
    \item We introduce a way for test set generation from Scannet++ by leveraging their noisy iPhone video data, for assessing blur, pose, and color drift.
\end{itemize}

\section{Related work}

Neural rendering was first proposed in NeRF~\cite{nerf}. It proposes to learn novel view synthesis by optimizing a radiance field representation so that it can re-render the training views. Many follow-up articles focused on the 3D representation used: either using an Multi Layer Perceptron (MLP)~\cite{fourierft,mipnerf,mipnerf360} voxel grids~\cite{plenoxels}, factorized voxel grids~\cite{tensorf,hexplane} and hash grids~\cite{instantngp,zipnerf}. Such method consistently showed increased speed both in training and rendering. Recently, Gaussian Splatting~\cite{kerbl3Dgaussians} introduced a new paradigm for neural rendering. Instead of optimizing a radiance field by querying points in 3D, it encodes the scene as a collection of 3D gaussians that can be splatted on the image plane for rendering. It has shown state-of-the-art rendering quality and real-time rendering. Neural rendering methods typically assume that the input data has good quality: pixel perfect camera poses, color consistency across views and sharp images. If those assumptions are not met, they produce bad quality results with typically blurry textures and floaters in the images. 

Color inconsistency was tackled in NeRF in the wild~\cite{nerfw} that deals with challenging data from raw internet image collections. It optimizes a per-image feature vector so that the color output also depend on the training image. This approach is used in many recent neural rendering methods. Another approach is to model the physical image capturing process as in~\cite{rawnerf,hdrnerf}. 

Pose inaccuracies was tackled in~\cite{nerfmm,barf,scnerf,camp} where the camera poses are optimized along the 3D representation. These papers introduce numerous techniques for optimizing noisy camera poses with a large noise, sometimes even random initialization. The main goal of such papers is to remove the dependency on Structure-from-Motion (SfM) step used before the actual 3D representation training. In this paper we also optimize poses but the objective is different: we assume a good first guess of the poses from SfM~\cite{colmap} and refine the poses during training.

Deblurring is a long standing problem in computer vision. Blind deblurring is the task of removing blur that was caused by an unknown convolution kernel. Existing work is splitted in single view deblurring and multi view. Single view deblurring is classically solved with optimization methods. Since the problem is ill-posed with infinite many solutions, classical methods heavily rely on regularization~\cite{chan1998total,rudin1992nonlinear}. With the advent of Deep Learning many papers proposed to apply it to deblurring~\cite{Tang_2019_CVPR,tao2018scale,Zamir_2022_CVPR,ren2020neural,ulyanov2018deep}. Multi-view deblurring relies on matching between multiple views~\cite{nah2019recurrent,chan2022basicvsr++} to aggregate informations. 

Neural rendering has provided an elegant way to do multi-view deblurring. The blur phenomenon can be directly modeled along with the rendering. DeblurNerf~\cite{deblurnerf} introduced a simple approach to train neural rendering models from blurry images. It relies on generating multiple rendering for a same training pixel, each rendering being produced from a slightly different ray as predicted by a small MLP. This idea was further improved with other architecture for offset prediction in~\cite{dpnerf,pdrf}. While the same idea could be applied to Gaussian Splatting, \cite{deblurgs} introduces a more efficient way to model blur for gaussian splatting models. It models blur with a new covariance matrix for each gaussians as predicted by an MLP. Our approach is similar to that approach but it explicitly derive the covariance changes from motion and defocus blur using a physical model. All deblurring methods that use neural rendering have shown impressive results on the DeblurNerf dataset~\cite{deblurnerf} but, to the best of our knowledge, they were never applied to data closer to the real world. We propose in this paper an evaluation on real-world data like Scannet++\cite{scannetpp}. 


\section{Brief review of Gaussian splatting}
Gaussian splatting is a scene reconstruction method introduced in ~\cite{kerbl3Dgaussians} that builds on top of ideas from~\cite{zwickerVolSplatting}. The underlying scene representation is non-parametric and consists in a collection of 3D Gaussian primitives $\Gamma\coloneqq\{\gamma_1,\ldots,\gamma_K\}$, which are rendered via volume splatting.

\paragraph{Gaussian primitive.}
A Gaussian primitive $\gamma_k$ can be imagined as a 3D Gaussian kernel
\[
\mathcal G_k(\vct x)\coloneqq \exp\left(-\frac{1}{2}(\vct x-\vct\mu_k)^\top\mat\Sigma_k^{-1}(\vct x-\vct\mu_k)\right)
\]
with mean $\vct\mu_k\in\mathbb R^3$ and covariance $\mat\Sigma_k$. Each primitive has an associated 
opacity factor $\alpha_k\in[0,1]$ and 
a feature vector $\vct f_k\in\mathbb R^d$. The feature vector typically holds spherical harmonics coefficients that encode view-dependent color information.

\paragraph{Splatting.} 
The operation of projecting a Gaussian primitive $\gamma_k$ to a camera pixel space is called \emph{splatting}. Let $\pi$ be the camera and assume the same notation to refer to its world-to-image transformation. The "splatted" primitive is approximated as a 2D Gaussian kernel $\mathcal G^\pi_k$ with mean $\vct\mu_k^\pi\coloneqq\pi(\vct\mu_k)$ and covariance $\mat\Sigma_k^\pi\coloneqq\mat J_k^{\pi}\mat\Sigma_k \mat {J_k^{\pi}}^\top$, where $\mat J_k^{\pi}$ denotes the Jacobian of $\pi$ evaluated at $\vct\mu_k$. Indeed, $\mathcal G_k^\pi$ is the kernel of the normal distribution we obtain by transforming a normal variate 3D point $\vct x\sim\mathcal G_k$ with $\pi$ approximated to the first order at the Gaussian primitive's center.

\paragraph{Rendering.} 
In order to render a scene represented as a collection of Gaussian primitives $\Gamma$ from a given camera $\pi$, we assume to have a decoder $\Phi$ that extracts the information we want to render from a primitive and optionally the pixel value (\eg~to get its associated view direction). The typical decoder that is used in Gaussian splatting denoted by $\Phi_\mathtt{RGB}$ turns spherical harmonics coefficients stored in $\vct f_k$ and the viewing direction associated to pixel $\vct u$ into an actual RGB color. The actual rendering for each pixel $\vct u$ consists in alpha-composing each primitives' decoded feature with alpha values given by $\alpha_k\mathcal G_k^\pi(\vct u)$. The order of the primitives in the alpha-composition is determined by the depth of their center when viewed from the target camera. We refer to~\cite{kerbl3Dgaussians} for more details.



\section{Improving Robustness of Gaussian splatting}
\label{sec:method}

\begin{figure}[t]
    \centering
    \includegraphics[width=0.9\textwidth]{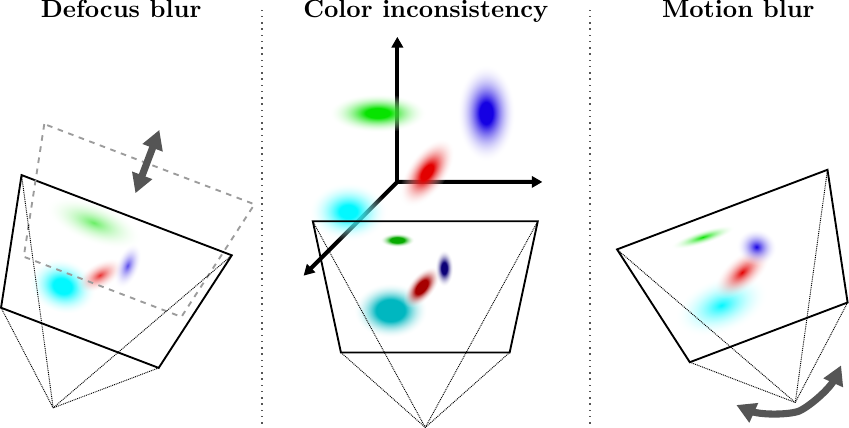}
    \caption{In 3DGS, defocus blur (left), color inconsistencies (center) and motion blur (right) can be modeled as simple transformations applied to the 3D Gaussian primitives (Sec.~\ref{sec:method}). This allows us to estimate per-camera motion, appearance and focus parameters, which can be factored out to recover a sharp reconstruction.}
    \label{fig:method}
\end{figure}

In this section we show how we can render Gaussian splatting robust against errors in the estimated camera poses (\emph{pose error}), against blurred images due to parts being out-of-focus (\emph{defocus blur}) or due to a camera being moved too fast while capturing (\emph{motion blur}), and against \emph{color inconsistencies} when observing the same part of the scene, \eg~due different white balancing, exposure time,~\etc.

\subsection{Robustness against pose errors and motion blur}
We start addressing two problems at once, namely how to correct errors in the camera poses used to fit a scene, and how to cope with images that are blurred due to camera motion.
In fact, the former problem is a special case of the latter. An image affected by motion blur can be regarded as the result of integrating images of a scene taken from multiple perturbed camera poses. As a special case, if the perturbed camera is only one (or multiple concentrated in the same point), this integration process yields an image that will be associated to a wrong pose, which is an instance of the pose correction problem. The implication of this is that we can formulate a solution to the motion blur problem and expect this to also cope with errors in the provided poses.

To solve the motion blur problem we need to estimate a distribution of camera poses whose renderings yield the ground-truth observation once integrated. 
An alternative, but equivalent, way of seeing the problem is to find a distribution of world transformations, while retaining the original cameras.
Indeed, each random world transformation changes the world reference frame in a way that renderings produced according to the original camera parameters are equivalent to renderings taken from a different camera in the original world reference frame.
We indirectly model a distribution of world transformations by constructing a function $\zeta[\theta_\pi^\mathtt{MB}](\vct x, \vct\epsilon)$ that depends on image-specific parameters $\theta_\pi^\mathtt{MB}$ and on a noise variable $\vct \epsilon$, and that implements a rigid transformation of $\vct x$ if we fix the other variables.

By exploiting the same idea underlying the splatting operation, we take the first-order approximation of $\zeta$ in $(\vct\mu_k,\vct 0)$ to get a linear function in $\vct x$ and $\vct \epsilon$. Assuming $\vct x\sim\mathcal G_k$ and $\vct \epsilon$ a standard normal multivariate, the output of the approximated transformation is Gaussian distributed with mean $\vct\mu_k^\zeta\coloneqq\zeta(\vct\mu_k, \vct 0)$ and covariance $\mat\Sigma_k^\zeta\coloneqq\mat J_{\vct x,k}^\zeta\mat\Sigma_k{\mat J_{\vct x,k}^{\zeta\top}}+\mat J_{\vct \epsilon,k}^\zeta {\mat J_{\vct \epsilon,k}^{\zeta\top}}$, where $\mat J_{\vct x,k}^\zeta$ is the Jacobian of $\zeta$ with respect to $\vct x$ evaluated at $(\vct\mu_k,\vct 0)$, and similarly for $\vct\epsilon$. As a result, we obtain updated 3D Gaussian primitives that integrate already the effect of the motion blur and can now be splatted and rendered as usual from the original camera poses.  

Finally, if we consider the 3D Gaussian kernel $\mathcal G_k^\zeta$ of the updated primitives, the overall opacity mass has increased due to the increase in the covariance. To compensate for this, we change also the opacity $\alpha_k$ of the original primitive, by scaling it by $\sqrt{\text{det}(\mat\Sigma_k)/\text{det}(\mat\Sigma^\zeta_k)}$.

\paragraph{Implementation details.} 


For the experiments in this paper, we model the world-transformation function $\zeta$ as follows:
\[
\zeta[\theta_\pi^\mathtt{MB}](\vct x, \vct\epsilon)\coloneqq \mat R_\pi\left(\exp\left(\mat\Sigma_\pi^{\mat R\frac{1}{2}}\vct\epsilon_{\mat R}\right)\vct x+\mat\Sigma_\pi^{t\frac{1}{2}}\vct\epsilon_t\right)+\vct t_\pi\,,
\]
where $\vct\epsilon\coloneqq(\vct\epsilon_{\mathtt R},\vct\epsilon_t)$, both being distributed as a standard 3D Gaussian, $\exp()$ is the exponential map, and $\theta_\pi^\mathtt{MB}\coloneqq(\mat R_\pi,\vct t_\pi, \mat\Sigma_\pi^{\mathtt R},\mat\Sigma_\pi^{t})$ are per-image parameters entailing a rotation $\mat R_\pi$, a translation $\vct t_\pi$, a covariance matrix $\mat\Sigma_\pi^{\mathtt R}$ for the rotation axis-angle representation and a covariance matrix $\mat\Sigma_\pi^t$ for the translation. Intuitively, $\zeta$ first remaps $\vct x$ according to a random rigid transformation, where the axis-angle representation of the rotation and the translation follow centered Gaussian distributions with covariance $\mat\Sigma_\pi^{\mathtt R}$ and $\mat\Sigma_\pi^t$, respectively. The result is then transformed via the rigid transformation that has $\mat R_\pi$ as the rotation matrix and $\vct t_\pi$ as the translation.
The new position and covariance of the 3D Gaussian primitives, considering the $\zeta$ function above are given by:
\[
\vct\mu_k^\zeta\coloneqq \mat R_\pi\vct\mu_k+\vct t_\pi\,,
\]
\[
\mat\Sigma_k^\zeta\coloneqq \mat R_\pi \left(\mat\Sigma_k+[\vct\mu_k]_\times \mat \Sigma_\pi^{\mathtt R}[\vct\mu_k]_\times^\top + \mat\Sigma_\pi^t\right)\mat R_\pi^\top\,,
\]
where $[\cdot]_\times$ gives the skew-symmetric cross product matrix for the provided vector.

\subsection{Robustness against defocus blur}

Defocus blur is caused by the use of a real camera with a non-zero aperture. Gaussian splatting assumes to have a pinhole camera so that every 3D point will be projected to a 2D point on the image. Unfortunately, this assumption is not true when the camera has a significant aperture, which happens especially with indoor scenes where the light is low or in video captures  where the exposure time needs to be small. When the aperture size is not negligible, only points that are on a 3D plane called \textit{focus plane} are in focus. The other points are projected to a circle that gets wider as we depart from the focus plane. The radius $R^\pi_k$ of this circle for a Gaussian primitive $\gamma_k$ and camera $\pi$ can be written as
\begin{equation}\label{eq:radius}
    R^\pi_k \coloneqq A_\pi\left(\rho_\pi-\frac{1}{D^\pi_k}\right)\,,
\end{equation}
where
\[
A_\pi\coloneqq\frac{a_\pi}{1-\rho_\pi f_\pi}\,.
\]
Here, $a_\pi$ denotes aperture of camera $\pi$, $f_\pi$ denotes its focal length, $\rho_\pi$ the inverse depth of its focus plane and $D_k^\pi$ is the depth of $\vct\mu_k$ when viewed from camera $\pi$'s perspective.

In order to address defocus blur in Gaussian splatting the idea is to add an offset corresponding to ${R^\pi_k}^2$ to the diagonal of the 2D covariance matrices $\mat \Sigma_k^\pi$ of each Gaussian primitive $\gamma_k$.
Since not all variables of the physical model are available (\eg aperture and focus plane distance are usually not known), we introduce image-specific variables, namely $\theta^\mathtt{DB}_\pi\coloneqq(A_\pi,\rho_\pi)$ that appear in \cref{eq:radius}, and optimize them directly during scene fitting.

Akin to what we have done for the motion blur case, we rescale the opacity $\alpha_k$ of the original primitive to compensate for the change in total opacity mass due to the added covarance terms.

\subsection{Robustness against color inconsistencies}
There are many possible factors that induce an inconsistency in the color we observe in two views of the same part of the scene. Those could be external, like sun suddenly being occluded by cloud between two shots, or camera-related,  \eg a change in the white balancing setting of the camera. These inconsistencies might severely harm the final reconstruction, often leading to floaters. To address this problem we introduce a new RGB decoder $\Psi_\mathtt{RGB}[\theta^\mathtt{CI}_\pi](\gamma_\pi,\vct u)$ that depends on additional per-image parameters $\theta^\texttt{CI}_\pi$. This gives the model the flexibility of correcting the effect of image-specific color nuisances. In this paper, we explore the simple solution of considering $\theta^\mathtt{CI}_\pi\coloneqq(\mat W_\pi,\vct q_\pi)$ the parameters of an affine color transformation that we apply to the standard decoded RGB color, \ie $\Psi_\mathtt{RGB}[\theta^\texttt{CI}_\pi](\gamma_\pi,\vct u)\coloneq \mat W_\pi\Phi_\mathtt{RGB}(\gamma_k,\vct u)+\vct q_\pi$. The advantage of this choice is that the average transformation that we apply at test time can be absorbed into a modification of the spherical harmonics stored in $\vct f_k$, making the resulting representation compatible with available Gaussian splatting viewers.

\subsection{Summary}
We provide in \cref{alg:training} the pseudocode of the rendering process at training time. 

\begin{algorithm}
\caption{Overview of the rendering process at train time}
\hspace*{\algorithmicindent} \textbf{Input:} Gaussians $\gamma_k \coloneqq (\vct\mu_k, \mat\Sigma_k, \alpha_k, \vct f_k)$, camera $\pi$\newline 
\hspace*{\algorithmicindent} per-image parameters $\mat R_\pi, \vct t_\pi, \mat\Sigma^\mathtt{R}_\pi,\mat\Sigma^t_\pi, A_\pi, \rho_\pi, \theta_\pi^\mathtt{CI}$ \\
\begin{algorithmic}
\For{k in $1 \dots K$}
    \State $\mat\Sigma_k^\zeta\gets \mat R_\pi\left(\mat\Sigma_k + \mat\Sigma^t_\pi + [\vct\mu_k]_{\times}\mat\Sigma^{\mathtt R}_\pi[\vct\mu_k]_{\times}^\top\right)\mat R_\pi^\top$ \Comment{Motion blur updated covariance}
    \State $\vct\mu_k^\zeta\gets \mat R_\pi\vct\mu_k+\vct t_\pi\,,$ \Comment{Motion blur updated position}
    \State $\vct\mu_k^\pi,\mat\Sigma_k^\pi\gets\texttt{splatting}(\gamma_k, \pi)$ \Comment{Project Gaussians on image}
    \State $D_k^\pi\gets\texttt{get\_depth}(\vct\mu_k^\pi)$ \Comment{Get primitive's depth}
    \State $\mat{\Sigma_k^{\pi}}' \gets \mat\Sigma_k^{\pi} + A_\pi^2(\rho_\pi - 1 / D_k^\pi)^2 \mat I$ \Comment{Apply defocus blur on image plane}
    \State $\alpha_{k} \gets \alpha_{k}\sqrt{\frac{\text{det}(\mat\Sigma_k)}{\text{det}(\mat\Sigma_k^\zeta)}}\sqrt{\frac{\text{det}(\mat\Sigma^\pi_k)}{\text{det}(\mat{\Sigma^\pi_k}')}}$ \Comment{Opacity scaling}
    \State $\mat\Sigma_k^{\pi} \gets \mat{\Sigma_k^{\pi}}'$
\EndFor
\State \textbf{return} \texttt{rendering}$(\Gamma, \Psi_\texttt{RGB}[\theta_\pi^\mathtt{CI}])$  \Comment{Rendering with color transformation}
\end{algorithmic}
\label{alg:training}
\end{algorithm}

\section{Experiments}

While training from noisy data is challenging, evaluating such models can also be hard.
In Sec.~\ref{sec:noisy_eval} we propose an evaluation procedure aimed at ensuring fair comparisons when only noisy data is available for testing.
Following this, in Sec.~\ref{sec:main_results} we report results on real world scenes from the Scannet++ dataset~\cite{scannetpp}, and in Sec.~\ref{sec:deblurnerf} we evaluate on the synthetic Deblurnerf~\cite{deblurnerf} dataset.
Finally, in Sec.~\ref{sec:ablations} we present a detailled ablation study of our contributions.

\subsection{Evaluating NVS from noisy captures}
\label{sec:noisy_eval}

In order to compare several baselines on real-world data as opposed to synthetic datasets, we would ideally require challenging training views, and corresponding sets of high-quality, clean test views.
The ScanNet++ dataset~\cite{scannetpp} was constructed following this logic: for each scene, it contains both a low quality smartphone video, and a set of images taken with a high end DSLR camera.
Unfortunately, the smartphone captures do not generally cover all scene regions observed by the DSLR images.
Using the high-quality data as a test set would thus mostly evaluate a model's ability to correctly ``guess'' the appearance of unobserved regions.
We therefore choose to use ScanNet++~\cite{scannetpp} smartphone videos only, which solves the scene coverage problem but implies we'll have to deal with a certain amount of noise in the test set.

 \begin{figure}[h]
    \centering

    \begin{subfigure}[t]{0.45\linewidth}
        \includegraphics[width=\linewidth]{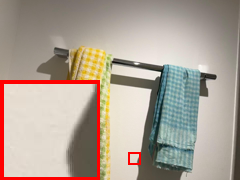}
        \caption{Evaluation image}
    \end{subfigure} \\
    \begin{subfigure}[t]{0.45\linewidth}
        \includegraphics[width=\linewidth]{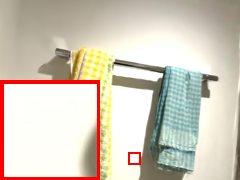}
        \caption{Raw rendering PSNR=14.82}
        \label{fig:raw_render}
    \end{subfigure}
    \hfill
    \begin{subfigure}[t]{0.45\linewidth}
        \includegraphics[width=\linewidth]{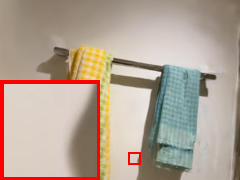}
        \caption{Test time optim. PSNR=25.85}
        \label{fig:opt_render}
    \end{subfigure}
    \caption{We perform test-time optimization of the per-image parameters to ensure a fair comparison. Without it, the renderings are noticeably misaligned and they have a color shift compared to the ground truth.}
    \label{fig:example-tt}
\end{figure}

First, the testing views may be blurred.
Second, images may be localized with poor accuracy, since the SfM algorithm (namely, COLMAP~\cite{colmap}) used by the authors of~\cite{scannetpp} to reconstruct the sequences is also negatively impacted by poor image quality.
Another source of pose uncertainty is the fact that we use pose optimization during training, meaning that our reconstructed world frame might drift over time, a problem known as \textit{gauge freedom} in the SfM literature.
Finally, colors might shift throughout the captures, as the smartphone's camera dynamically adapts exposure and white balance. 
The last two points imply that naively rendering a trained model from the given test camera poses will likely produce images that don't match the ground truth, as illustrated in Fig.~\ref{fig:raw_render}. 

To address blur, we select test views based on a blurriness measure. 
Specifically, and for each scene, we sort its images by decreasing maximum gradient magnitude, then iteratively select the 10 best while ensuring no test view is within $0.5$m and $60^\circ$ of another.
To address the issues related to pose and color drift, we propose to do a test-time adaptation to align the camera poses and the color profile to the expected one from ground truth, before computing the evaluation metrics.
This is shown in~\ref{fig:opt_render} where the camera poses and the color transformation were optimized for 1000 steps.
Given the small number of optimized parameters we argue that this approach does not lead to over-fitting, and that it is a fair comparison since we perform this optimization for all compared baselines.


\subsection{Comparison with NVS baselines on real-world data}
\label{sec:main_results}

Following the procedure described in the previous section, we build a real-world benchmark including 49 scenes from the ScanNet++ validation set, and compare against NVS baselines.
Specifically, we consider Mip-NeRF 360~\cite{mipnerf360} with per-image appearance codes~\cite{nerfw} as a representative of state-of-the-art NeRF approaches, along with Nerfacto~\cite{nerfstudio} and standard Gaussian Splatting~\cite{kerbl3Dgaussians}.
To ensure a fair comparison, we augment all models with pose optimization capabilities, by simply enabling gradient propagation to the pose parameters.
During test-time adaptation, we freeze all model parameters except the pose offsets and the per-image color parameters.
Note that, in the case of Mip-NeRF 360 and Nerfacto, this might lead to greater over-fitting than with our model, since the per-image codes, while low-dimensional, are used as inputs to a highly non-linear function that might produce appearance changes beyond simple color adaptation. We therefore compare with an additional adaptation that does not use per-image codes but an affine color transformation similar to our approach.
As shown in Tab.~\ref{tab:sota} Nerfacto produces the best results for the PSNR metrics. This is probably because it is more strongly regularized and it can produce better uniformly colored regions. However, our approach outperforms the baselines on the perceptual metrics (SSIM and LPIPS) that are more sensitive to fine grained details.

Additionally, we augment Mip-NeRF 360 and Nerfacto with a deblurring component, adapted from the formulation of DeblurNerf~\cite{deblurnerf}.
Interestingly, this approach noticeably decreases accuracy compared to standard models.
We suspect this might be related to the fact that DeblurNerf drastically increases the training-time memory requirements, forcing us to reduce batch size by a factor $5$ compared to the value proposed in the original papers and used in the other experiments.
It is also possible that the architecture might already be close to the limits of its capacity when confronted with the large and highly-detailed scenes in ScanNet++, independently of whether or not specific steps are taken to contrast dataset noise.



\begin{table}[t]
    \centering
    \caption{Comparison with state of the art approaches. Note that all models are trained with per-image color adaptation and pose refinement.}
    \label{tab:sota}
    \begin{tabular}{l|ccc}
    \toprule
        Method & PSNR$\uparrow$ & SSIM$\uparrow$ & LPIPS$\downarrow$ \\
    \midrule
        Mipnerf360~\cite{mipnerf360} + image codes & 23.90 & .8340 & .3634 \\
        Mipnerf360~\cite{mipnerf360} + image codes + deblurNerf~\cite{deblurnerf} & 23.43 & .8172 & .4084 \\
        Mipnerf360~\cite{mipnerf360} + affine color transformation & 25.61 & .8406 & .3677 \\
        \midrule
        Nerfacto~\cite{nerfstudio} + image codes  & \bf 27.09 & .8486 & .3571 \\
        Nerfacto~\cite{nerfstudio} + image codes  + deblurNerf~\cite{deblurnerf} & 26.71 & .8392 & .3793 \\
        Nerfacto~\cite{nerfstudio} + affine color transformation & 26.74 & .8389 & .4070 \\
        \midrule
        Gaussian splatting~\cite{kerbl3Dgaussians} &  23.87 & .8434 & .3493  \\
    \midrule
        Ours  & 24.08 & \bf .8503 & \bf .3419 \\
    \bottomrule
    \end{tabular}
\end{table} 

\begin{figure}[htp!]
    \centering
    \begin{subfigure}[t]{0.24\textwidth}
        \includegraphics[width=\textwidth]{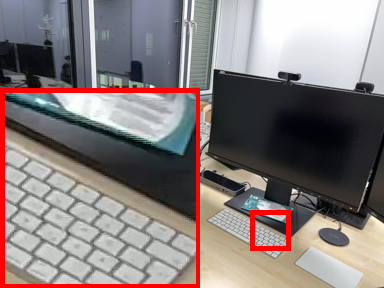}
    \end{subfigure}
    \begin{subfigure}[t]{0.24\textwidth}
        \includegraphics[width=\textwidth]{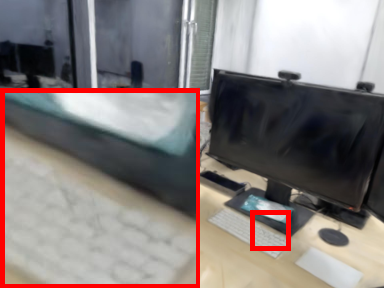}
    \end{subfigure}
    \begin{subfigure}[t]{0.24\textwidth}
        \includegraphics[width=\textwidth]{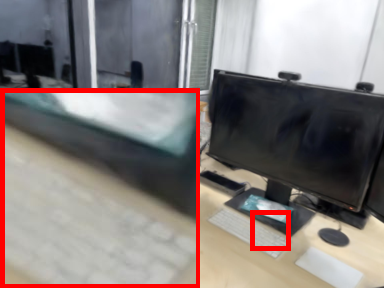}
    \end{subfigure}
    \begin{subfigure}[t]{0.24\textwidth}
        \includegraphics[width=\textwidth]{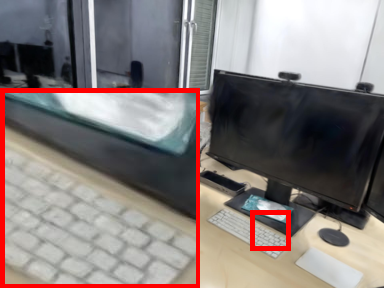}
    \end{subfigure}

    \begin{subfigure}[t]{0.24\textwidth}
        \includegraphics[width=\textwidth]{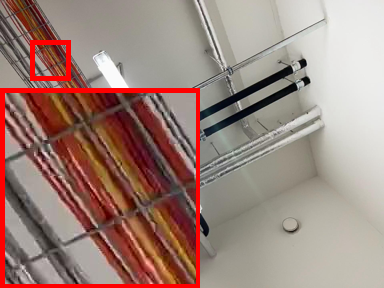}
    \end{subfigure}
    \begin{subfigure}[t]{0.24\textwidth}
        \includegraphics[width=\textwidth]{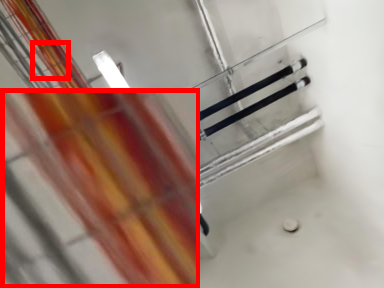}
    \end{subfigure}
    \begin{subfigure}[t]{0.24\textwidth}
        \includegraphics[width=\textwidth]{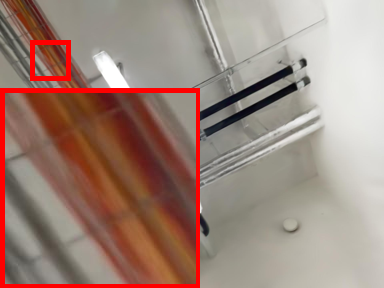}
    \end{subfigure}
    \begin{subfigure}[t]{0.24\textwidth}
        \includegraphics[width=\textwidth]{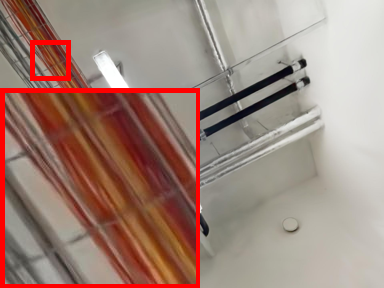}
    \end{subfigure}

    \begin{subfigure}[t]{0.24\textwidth}
        \includegraphics[width=\textwidth]{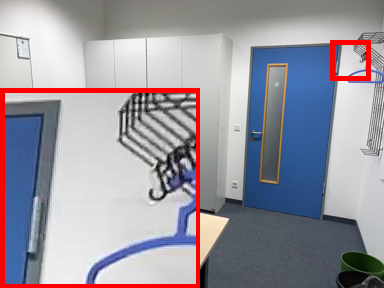}
    \end{subfigure}
    \begin{subfigure}[t]{0.24\textwidth}
        \includegraphics[width=\textwidth]{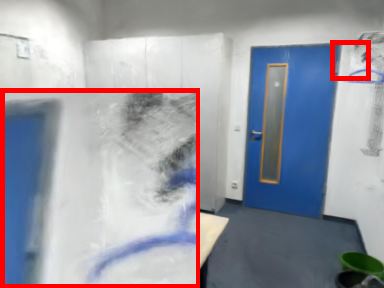}
    \end{subfigure}
    \begin{subfigure}[t]{0.24\textwidth}
        \includegraphics[width=\textwidth]{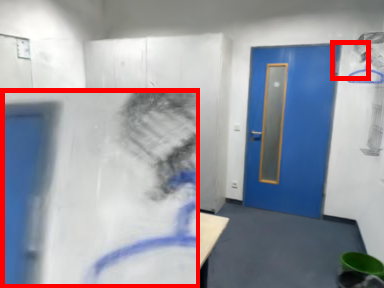}
    \end{subfigure}
    \begin{subfigure}[t]{0.24\textwidth}
        \includegraphics[width=\textwidth]{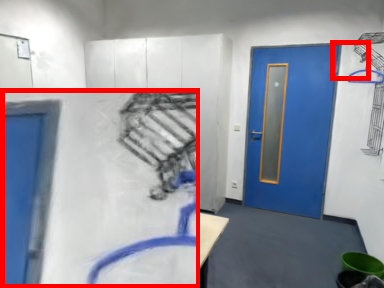}
    \end{subfigure} 

    \begin{subfigure}[t]{0.24\textwidth}
        \includegraphics[width=\textwidth]{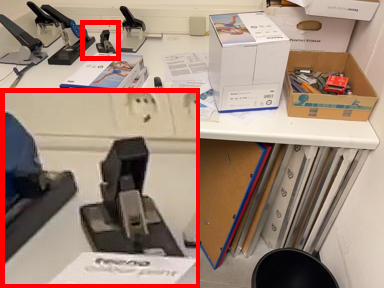}
    \end{subfigure}
    \begin{subfigure}[t]{0.24\textwidth}
        \includegraphics[width=\textwidth]{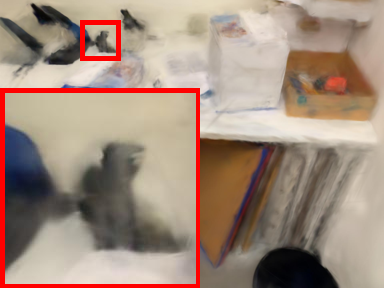}
    \end{subfigure}
    \begin{subfigure}[t]{0.24\textwidth}
        \includegraphics[width=\textwidth]{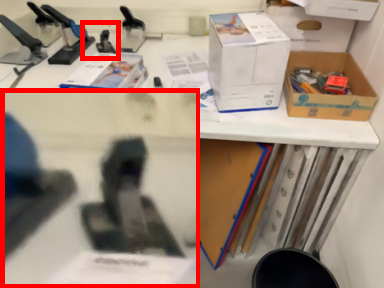}
    \end{subfigure}
    \begin{subfigure}[t]{0.24\textwidth}
        \includegraphics[width=\textwidth]{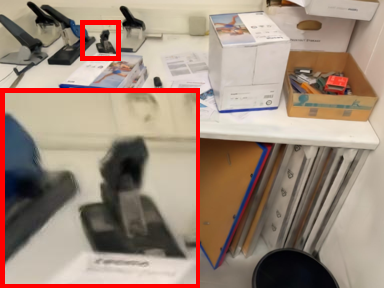}
    \end{subfigure}

    \begin{subfigure}[t]{0.24\textwidth}
        \includegraphics[width=\textwidth]{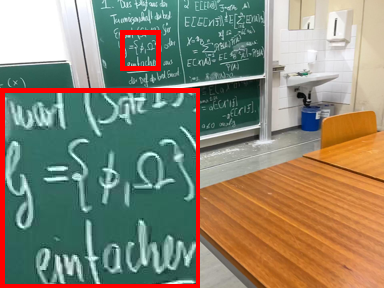}
    \end{subfigure}
    \begin{subfigure}[t]{0.24\textwidth}
        \includegraphics[width=\textwidth]{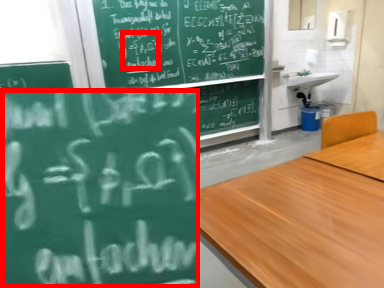}
    \end{subfigure}
    \begin{subfigure}[t]{0.24\textwidth}
        \includegraphics[width=\textwidth]{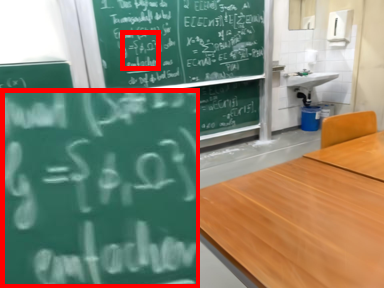}
    \end{subfigure}
    \begin{subfigure}[t]{0.24\textwidth}
        \includegraphics[width=\textwidth]{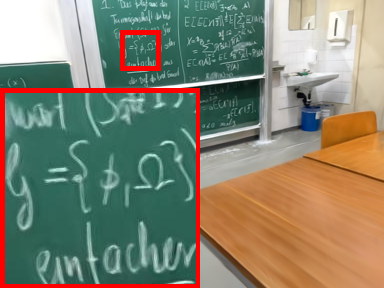}
    \end{subfigure}

    \begin{subfigure}[t]{0.24\textwidth}
        \includegraphics[width=\textwidth]{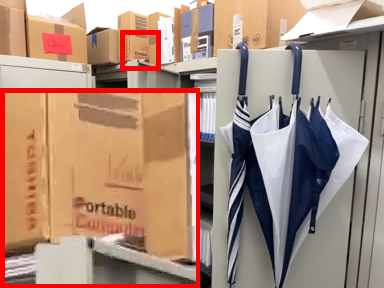}
        \caption{Ground truth}
    \end{subfigure}
    \begin{subfigure}[t]{0.24\textwidth}
        \includegraphics[width=\textwidth]{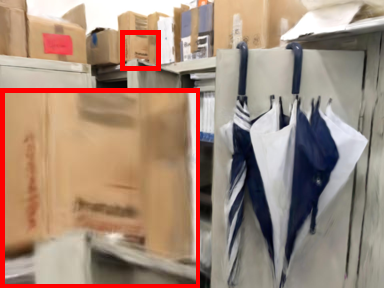}
        \caption{Gaussian splatting}
    \end{subfigure}
    \begin{subfigure}[t]{0.24\textwidth}
        \includegraphics[width=\textwidth]{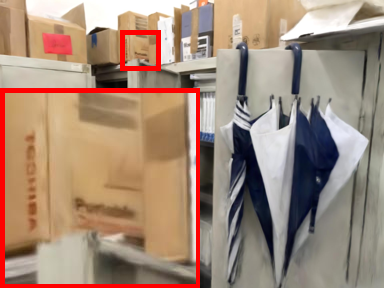}
        \caption{GS with pose opt.}
    \end{subfigure}
    \begin{subfigure}[t]{0.24\textwidth}
        \includegraphics[width=\textwidth]{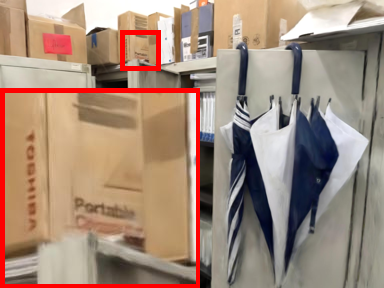}
        \caption{Ours}
    \end{subfigure}

    \begin{subfigure}[t]{0.24\textwidth}
        \includegraphics[width=\textwidth]{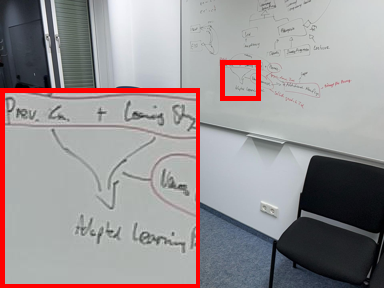}
        \caption{Ground truth}
    \end{subfigure}
    \begin{subfigure}[t]{0.24\textwidth}
        \includegraphics[width=\textwidth]{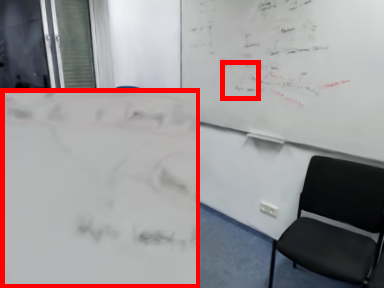}
        \caption{Gaussian splatting}
    \end{subfigure}
    \begin{subfigure}[t]{0.24\textwidth}
        \includegraphics[width=\textwidth]{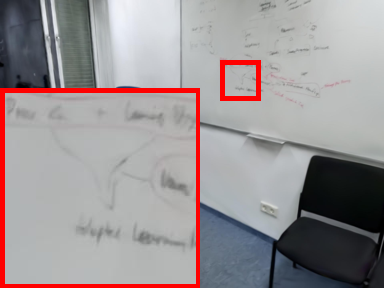}
        \caption{GS with pose opt.}
    \end{subfigure}
    \begin{subfigure}[t]{0.24\textwidth}
        \includegraphics[width=\textwidth]{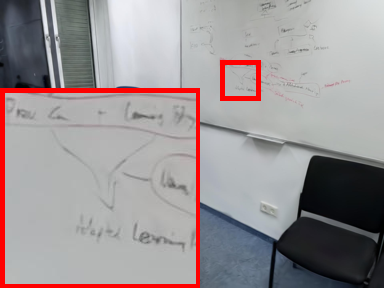}
        \caption{Ours}
    \end{subfigure}

    \caption{Qualitative comparison of GS models with different ablations.}
    \label{fig:ablation}
\end{figure}

\subsection{Ablation study}
\label{sec:ablations}

We use the evaluation procedure described in Sec.~\ref{sec:noisy_eval} to evaluate the relative contributions of the component we describe in Sec.~\ref{sec:method}.
In particular, we separately consider affine color transformation (Color transfo.), pose optimization (Pose optim.), motion blur modeling (Motion) and defocus blur modeling (Defocus).
Results are reported in Table~\ref{tab:ablation}. 

\begin{table}[t]
    \centering
    \caption{Ablation study demonstrating proposed robustness mechanism modules.}
    \label{tab:ablation}
    \begin{tabular}{c c c c|c c c}
        \toprule
        
        Color transfo. & Pose optim. & Motion & Defocus & PSNR$\uparrow$ & SSIM$\uparrow$ & LPIPS$\downarrow$ \\
        \midrule
         & & &  & 21.71& .807 & .395\\
        \checkmark & & &  & 23.08 & .817 & .380 \\
        \checkmark & \checkmark & & & 23.87 & .843 & .349 \\
        \checkmark & \checkmark & \checkmark &  & \bf 24.08 & \bf .850 & \bf .341 \\
        \checkmark & \checkmark & & \checkmark & 23.94 & .846 & .346\\
        \checkmark & \checkmark & \checkmark & \checkmark & \bf 24.08 & \bf .850 & .342 \\ 
        \bottomrule
    \end{tabular}
\end{table}

The simple additions of color transformation and pose optimization clearly show a large improvements over the baseline Gaussian Splatting.
Since removing color transformation clearly leads to severe performance degradation, we keep it enable for all other ablation experiments and to generate the qualitative results in Fig.~\ref{fig:ablation}.
Defocus blur modeling and motion blur modeling are both useful individually but the use of both simultaneously does not improve over motion blur modeling alone.
This is probably because there are few images with defocus blur in the ScanNet++ scenes we use for evaluation.
Interestingly, defocus blur modeling still improves over no blur modeling, since it can (inaccurately) explain part of the motion blur.
Defocus blur is however useful for DeblurNerf data~\cite{deblurnerf} as shown in Sec.~\ref{sec:deblurnerf}.

\subsection{Evaluation on synthetic data}
\label{sec:deblurnerf}

We finally perform experiments on the synthetic scenes of the DeblurNeRF~\cite{deblurnerf} dataset.
This dataset consists of multiple scenes with either motion or defocus blur. 
While this dataset is useful for benchmarking deblurring capabilities of models, it shows a quite different set of challenges compared to the casually captured smartphone videos of ScanNet++.
Specifically, the scenes are smaller and viewed from forward facing cameras, and the blur is significantly more pronounced.
On the other hand, the dataset provides a perfect setup where the color are consistent, poses are accurate and test images are perfectly sharp.
Due to this, we can evaluate without color correction, pose optimization or test-time adaptation.

We compare our method with state-of-the-art deblurring NVS methods, showing results in Table~\ref{tab:deblurnerf}.
While our method is significantly below current NeRF based state-of-the-art, it still significantly outperforms standard Gaussian Splatting for both motion and defocus blur.
We suspect this might be a direct consequence of the greater regularization imposed by the MLPs in the NeRFs, which allow them to deal with extreme levels of blur better than Gaussian Splatting-based approaches.

\begin{table}[h]
    \centering
    \caption{Results on Deblurnerf~\cite{deblurnerf}  synthetic dataset}
    \label{tab:deblurnerf}
    \begin{tabular}{l|cc|cc}
    \toprule
        \multirow{2}{*}{Method} & \multicolumn{2}{|c|}{Motion blur} & \multicolumn{2}{|c}{Defocus} \\
        & PSNR$\uparrow$ & SSIM$\uparrow$ & PSNR$\uparrow$ & SSIM$\uparrow$ \\
    \midrule
        NeRF~\cite{nerf} & 25.93 & .7791 & 25.83 & .7701 \\
        deblurNerf~\cite{deblurnerf} & 28.77 & .8593 & 28.37 & .8527\\
        dp-nerf~\cite{dpnerf} & 29.23 & .8674 & 29.33 & .8713\\
        PDRF-10~\cite{pdrf} & \bf 29.29 & \bf .8798 & \bf 30.08 & .8931 \\
    \midrule
        Gaussian Splatting~\cite{kerbl3Dgaussians} & 21.97 & .6808 & 24.46 & .8199 \\
        Deblur-GS~\cite{deblurgs} & - & - & 29.43 & .8907 \\
        Ours & 23.99 & .7811 & 29.24 & \bf .9168\\
    \bottomrule
    \end{tabular}
\end{table}

\section{Limitations}

While our paper makes a significant contribution toward real-world robust Gaussian splattings, it suffers from limitations. First we add modelisation for defocus blur and camera motion blur but we have not tackled other types of motion blur. Such blur could \eg come from non-static objects like walking pedestrians, swaying trees, or moving cars. Incorporating compensation strategies for dynamic blur would be much more challenging since it would also require us to handle dynamic scenes. Second, our method does not solve the poor generalization of 3DGS to viewpoints far away from the training trajectory. This is a well-known issue for 3DGS and our method keeps sharing that limitation. 

\section{Conclusions}
In this paper, we presented a new approach to improve the robustness of 3D Gaussian splats (3DGS) for practical applications involving real-world data challenges. We introduced a formulation that addresses common error sources such as blur, imperfect camera poses, and color inconsistencies. Our solution models motion blur as a Gaussian distribution over camera poses, allowing us to address both camera pose refinement and motion blur correction while maintaining the benefits of 3DGS in terms of training efficiency and rendering speed. Additionally, we propose a defocus blur compensation mechanism and an RGB decoder function with per-image parameters to address color inconsistencies caused by ambient light, shadows, or camera-specific issues. We experimentally validated our contributions on relevant benchmark datasets including Scannet++ and Deblur-NeRF, showing consistent improvements over relevant baselines.

\bibliographystyle{splncs04}
\bibliography{main}

\begin{thebibliography}{10}
\providecommand{\url}[1]{\texttt{#1}}
\providecommand{\urlprefix}{URL }
\providecommand{\doi}[1]{https://doi.org/#1}

\bibitem{mipnerf}
Barron, J.T., Mildenhall, B., Tancik, M., Hedman, P., Martin-Brualla, R.,
  Srinivasan, P.P.: Mip-nerf: A multiscale representation for anti-aliasing
  neural radiance fields. In: ICCV. pp. 5855--5864 (2021)

\bibitem{mipnerf360}
Barron, J.T., Mildenhall, B., Verbin, D., Srinivasan, P.P., Hedman, P.:
  Mip-nerf 360: Unbounded anti-aliased neural radiance fields. In: CVPR (2022)

\bibitem{zipnerf}
Barron, J.T., Mildenhall, B., Verbin, D., Srinivasan, P.P., Hedman, P.:
  Zip-nerf: Anti-aliased grid-based neural radiance fields (2023)

\bibitem{hexplane}
Cao, A., Johnson, J.: Hexplane: A fast representation for dynamic scenes. In:
  CVPR. pp. 130--141 (June 2023)

\bibitem{chan2022basicvsr++}
Chan, K.C., Zhou, S., Xu, X., Loy, C.C.: Basicvsr++: Improving video
  super-resolution with enhanced propagation and alignment. In: CVPR. pp.
  5972--5981 (2022)

\bibitem{chan1998total}
Chan, T.F., Wong, C.K.: Total variation blind deconvolution. IEEE TIP
  \textbf{7}(3),  370--375 (1998)

\bibitem{tensorf}
Chen, A., Xu, Z., Geiger, A., Yu, J., Su, H.: Tensorf: Tensorial radiance
  fields. In: ECCV (2022)

\bibitem{plenoxels}
Fridovich-Keil, S., Yu, A., Tancik, M., Chen, Q., Recht, B., Kanazawa, A.:
  Plenoxels: Radiance fields without neural networks. In: Proceedings of the
  IEEE/CVF Conference on Computer Vision and Pattern Recognition (CVPR). pp.
  5501--5510 (June 2022)

\bibitem{hdrnerf}
Huang, X., Zhang, Q., Feng, Y., Li, H., Wang, X., Wang, Q.: Hdr-nerf: High
  dynamic range neural radiance fields. In: CVPR. pp. 18398--18408 (2022)

\bibitem{scnerf}
Jeong, Y., Ahn, S., Choy, C., Anandkumar, A., Cho, M., Park, J.:
  Self-calibrating neural radiance fields. In: ICCV. pp. 5846--5854 (2021)

\bibitem{kerbl3Dgaussians}
Kerbl, B., Kopanas, G., Leimk{\"u}hler, T., Drettakis, G.: 3d gaussian
  splatting for real-time radiance field rendering. ACM Transactions on
  Graphics  \textbf{42}(4) (July 2023),
  \url{https://repo-sam.inria.fr/fungraph/3d-gaussian-splatting/}

\bibitem{deblurgs}
Lee, B., Lee, H., Sun, X., Ali, U., Park, E.: Deblurring 3d gaussian splatting.
  arXiv preprint arXiv:2401.00834  (2024)

\bibitem{dpnerf}
Lee, D., Lee, M., Shin, C., Lee, S.: Dp-nerf: Deblurred neural radiance field
  with physical scene priors. In: CVPR. pp. 12386--12396 (June 2023)

\bibitem{barf}
Lin, C.H., Ma, W.C., Torralba, A., Lucey, S.: Barf: Bundle-adjusting neural
  radiance fields. In: ICCV. pp. 5741--5751 (October 2021)

\bibitem{deblurnerf}
Ma, L., Li, X., Liao, J., Zhang, Q., Wang, X., Wang, J., Sander, P.V.:
  Deblur-nerf: Neural radiance fields from blurry images. In: CVPR. pp.
  12861--12870 (June 2022)

\bibitem{nerfw}
Martin-Brualla, R., Radwan, N., Sajjadi, M.S.M., Barron, J.T., Dosovitskiy, A.,
  Duckworth, D.: {NeRF in the Wild: Neural Radiance Fields for Unconstrained
  Photo Collections}. In: CVPR (2021)

\bibitem{rawnerf}
Mildenhall, B., Hedman, P., Martin-Brualla, R., Srinivasan, P.P., Barron, J.T.:
  Nerf in the dark: High dynamic range view synthesis from noisy raw images.
  In: CVPR. pp. 16190--16199 (2022)

\bibitem{nerf}
Mildenhall, B., Srinivasan, P.P., Tancik, M., Barron, J.T., Ramamoorthi, R.,
  Ng, R.: Nerf: Representing scenes as neural radiance fields for view
  synthesis. In: ECCV (2020)

\bibitem{instantngp}
M{\"u}ller, T., Evans, A., Schied, C., Keller, A.: Instant neural graphics
  primitives with a multiresolution hash encoding. ACM TOG  \textbf{41}(4),
  1--15 (2022)

\bibitem{nah2019recurrent}
Nah, S., Son, S., Lee, K.M.: Recurrent neural networks with intra-frame
  iterations for video deblurring. In: CVPR. pp. 8102--8111 (2019)

\bibitem{camp}
Park, K., Henzler, P., Mildenhall, B., Barron, J.T., Martin-Brualla, R.: Camp:
  Camera preconditioning for neural radiance fields. ACM TOG  (2023)

\bibitem{pdrf}
Peng, C., Chellappa, R.: Pdrf: progressively deblurring radiance field for fast
  scene reconstruction from blurry images. In: AAAI. vol.~37, pp. 2029--2037
  (2023)

\bibitem{ren2020neural}
Ren, D., Zhang, K., Wang, Q., Hu, Q., Zuo, W.: Neural blind deconvolution using
  deep priors. In: CVPR. pp. 3341--3350 (2020)

\bibitem{rudin1992nonlinear}
Rudin, L.I., Osher, S., Fatemi, E.: Nonlinear total variation based noise
  removal algorithms. Physica D: nonlinear phenomena  \textbf{60}(1-4),
  259--268 (1992)

\bibitem{colmap}
Sch\"{o}nberger, J.L., Frahm, J.M.: Structure-from-motion revisited. In: CVPR
  (2016)

\bibitem{fourierft}
Tancik, M., Srinivasan, P., Mildenhall, B., Fridovich-Keil, S., Raghavan, N.,
  Singhal, U., Ramamoorthi, R., Barron, J., Ng, R.: Fourier features let
  networks learn high frequency functions in low dimensional domains. NeurIPS
  \textbf{33},  7537--7547 (2020)

\bibitem{nerfstudio}
Tancik, M., Weber, E., Ng, E., Li, R., Yi, B., Kerr, J., Wang, T.,
  Kristoffersen, A., Austin, J., Salahi, K., Ahuja, A., McAllister, D.,
  Kanazawa, A.: Nerfstudio: A modular framework for neural radiance field
  development. In: ACM SIGGRAPH 2023 Conference Proceedings. SIGGRAPH '23
  (2023)

\bibitem{Tang_2019_CVPR}
Tang, C., Zhu, X., Liu, X., Wang, L., Zomaya, A.: Defusionnet: Defocus blur
  detection via recurrently fusing and refining multi-scale deep features. In:
  CVPR (June 2019)

\bibitem{tao2018scale}
Tao, X., Gao, H., Shen, X., Wang, J., Jia, J.: Scale-recurrent network for deep
  image deblurring. In: CVPR. pp. 8174--8182 (2018)

\bibitem{ulyanov2018deep}
Ulyanov, D., Vedaldi, A., Lempitsky, V.: Deep image prior. In: CVPR. pp.
  9446--9454 (2018)

\bibitem{nerfmm}
Wang, Z., Wu, S., Xie, W., Chen, M., Prisacariu, V.A.: Ne{RF}$--$: Neural
  radiance fields without known camera parameters. arXiv preprint
  arXiv:2102.07064  (2021)

\bibitem{scannetpp}
Yeshwanth, C., Liu, Y.C., Nie{\ss}ner, M., Dai, A.: Scannet++: A high-fidelity
  dataset of 3d indoor scenes. In: ICCV (2023)

\bibitem{Zamir_2022_CVPR}
Zamir, S.W., Arora, A., Khan, S., Hayat, M., Khan, F.S., Yang, M.H.: Restormer:
  Efficient transformer for high-resolution image restoration. In: CVPR. pp.
  5728--5739 (June 2022)

\bibitem{zwickerVolSplatting}
Zwicker, M., Pfister, H., van Baar, J., Gross, M.: Ewa volume splatting. In:
  Proceedings Visualization, 2001. VIS '01. pp. 29--538 (2001).
  \doi{10.1109/VISUAL.2001.964490}

\end{thebibliography}

\newpage

\appendix 

\section{Derivations for results in Subsection 4.1}

We show the derivations for the theoretical results of subsection 4.1. We first explain how to derive the updated mean and covariance of a gaussian transformed by a generic camera transformation. We then derive the updates used in practice with our noisy camera formulation. Finally we explain the opacity scaling used to keep the total mass constant when changing a gaussian's covariance.

\subsection{Parameters of updated 3D Gaussian primitives for generic $\zeta$}

The first-order approximation of $\zeta(\vct x,\vct\epsilon)$ at $(\vct\mu_k,\vct 0)$ is given by
\[
\tilde\zeta(\vct x,\vct \epsilon)\coloneqq \zeta(\vct \mu_k,\vct 0)+\underbrace{\mat J^\zeta_{\vct x}(\vct\mu_k,\vct 0)}_{\coloneqq\mat J^\zeta_{\vct x,k}}(\vct x-\vct\mu_k)+\underbrace{\mat J^\zeta_{\vct \epsilon}(\vct\mu_k,\vct 0)}_{\coloneqq\mat J^\zeta_{\vct \epsilon,k}}\vct\epsilon\,.
\]
Let $\vct x\sim\mathcal N(\vct\mu_k, \mat\Sigma_k)$ be normal distributed with mean $\vct\mu_k$ and covariance $\mat\Sigma_k$ and let $\vct\epsilon\sim\mathcal N(\vct 0, \mat I)$ be a standard normal multivariate. Then $\tilde\zeta(\vct x,\vct\epsilon)$ is also Gaussian distributed, for $\hat\zeta$ is linear is $(\vct x,\vct\epsilon)$, with mean and covariance being given by:
\begin{alignat*}{3}
\vct\mu_k^\zeta&\coloneqq\mathbb E[\tilde\zeta(\vct x,\vct\epsilon)]&=&\zeta(\vct\mu_k,\vct 0)\\
\mat\Sigma_k^\zeta&\coloneqq\mathbb V\text{ar}[\tilde\zeta(\vct x,\vct\epsilon)]&=&\mat J^\zeta_{\vct x,k}\mathbb V\text{ar}[\vct x-\vct\mu_k]\mat J^{\zeta\top}_{\vct x,k}&+&\mat J^\zeta_{\vct \epsilon,k}\mathbb V\text{ar}[\vct \epsilon]\mat J^{\zeta\top}_{\vct \epsilon,k}\\
&&=&\mat J^\zeta_{\vct x,k}\mat\Sigma_k\mat J^{\zeta\top}_{\vct x,k}&+&\mat J^\zeta_{\vct \epsilon,k}\mat J^{\zeta\top}_{\vct \epsilon,k}.
\end{alignat*}
\subsection{Opacity scaling for updated 3D Gaussian primitives}
The total opacity mass $\mathcal M_k$ of a 3D Gaussian primitive $\gamma_k$ is the integral of $\alpha_k\mathcal G_k(\vct x)$ over the Euclidean plane, which is given by:
\[
\mathcal M_k\coloneqq \alpha_k\int_{\mathbb R^2}\mathcal G_k(\vct x)d\vct x=\alpha_k\text{det}(2\pi\mat\Sigma_k)^\frac{1}{2}\int_{\mathbb R^2}\mathcal N(\vct x;\vct\mu_k,\mat\Sigma_k)d\vct x=2\pi\text{det}(\mat\Sigma_k)^\frac{1}{2}\alpha_k
\]
Similarly, we would have $\mathcal M_k^\zeta\coloneqq 2\pi\text{det}(\mat\Sigma^\zeta_k)^\frac{1}{2}\alpha^\zeta_k$. Now we want to find $\alpha_k^\zeta$ such that $\mathcal M_k=\mathcal M_k^\zeta$, which is given by
\[
\alpha^\zeta_k\coloneq\alpha_k\frac{2\pi\text{det}(\mat\Sigma_k)^\frac{1}{2}}{2\pi\text{det}(\mat\Sigma^\zeta_k)^\frac{1}{2}}=\alpha_k\sqrt{\frac{\text{det}(\mat\Sigma_k)}{\text{det}(\mat\Sigma^\zeta_k)}}\,.
\]
This shows where the proposed opacity scaling factor stems from.
\subsection{Parameters of updated 3D Gaussian primitives for specific $\zeta$}
Let's consider the following particular instance of $\zeta$:
\[
\zeta(\vct x,(\vct\epsilon_{\mat R}, \vct\epsilon_t))\coloneqq\mat R_\pi\left(\exp\left(\mat\Sigma_\pi^{\mat R\frac{1}{2}}\vct\epsilon_{\mat R}\right)\vct x+\mat\Sigma_\pi^{t\frac{1}{2}}\vct\epsilon_t\right)+\vct t_\pi\,.
\]
Then we have that:
\[
\vct\mu_k^\zeta\coloneqq\zeta(\vct\mu_k,(\vct 0, \vct 0))=\mat R_\pi\left(\underbrace{\exp\left(\mat\Sigma_\pi^{\mat R\frac{1}{2}}\vct 0\right)}_{\mat I}\vct \mu_k+\mat\Sigma_\pi^{t\frac{1}{2}}\vct 0\right)+\vct t_\pi=\mat R_\pi\vct \mu_k+\vct t_\pi\,.
\]
%
The Jacobians terms that are needed to compute the covariance are given by:
\[
\mat J^\zeta_{\vct x,k}\coloneqq \mat R_\pi\exp\left(\mat\Sigma_\pi^{\mat R\frac{1}{2}}\vct 0\right)=\mat R_\pi\,,\qquad\text{and}\qquad
\mat J^\zeta_{{\vct \epsilon},k}\coloneqq 
\begin{bmatrix}
\mat J^\zeta_{{\vct \epsilon_{\mat R}},k}&\mat J^\zeta_{{\vct \epsilon_t},k}
\end{bmatrix}\,,
\]
Let's write the first order approximation of $\zeta$ 
\wrt $\vct \epsilon_{\mat R}$

\begin{alignat*}{1}
    \zeta(\vct x,(\vct\epsilon_{\mat R}, \vct\epsilon_t))&\coloneqq\mat R_\pi\left(\exp\left(\mat\Sigma_\pi^{\mat R\frac{1}{2}}\vct\epsilon_{\mat R}\right)\vct x+\mat\Sigma_\pi^{t\frac{1}{2}}\vct\epsilon_t\right)+\vct t_\pi \\
    &\overset{(a)}{=} \mat R_\pi \left[\vct x+[\mat\Sigma_\pi^{\mat R\frac{1}{2}}\vct\epsilon_{\mat R}]_\times\vct x+ O(\Vert\vct\epsilon_{\mat R}\Vert^2) + \mat\Sigma_\pi^{t\frac{1}{2}}\vct\epsilon_t\right] + \vct t_\pi \\
    &\overset{(b)}{=} \mat R_\pi \left[\vct x - [\vct x]_\times\mat\Sigma_\pi^{\mat R\frac{1}{2}}\vct\epsilon_{\mat R}+ O(\Vert\vct\epsilon_{\mat R}\Vert^2) + \mat\Sigma_\pi^{t\frac{1}{2}}\vct\epsilon_t\right] + \vct t_\pi\,.
\end{alignat*}
Here, in $(a)$ we replace the exponential map $\exp(\vct z)$ with its Taylor expansion to the first-order at $\vct 0$, namely $\mat I+[\vct z]_\times+O(\Vert\vct z\Vert^2)$, and in (b) we use the property of the cross-product $[\vct a]_\times\vct b=-[\vct b]_\times\vct a$. 
We can now identify the jacobians expression from the above linear approximations:
\begin{alignat*}{1}
   \mat J^\zeta_{{\vct \epsilon}_t,k} &= \mat R_\pi\mat\Sigma_\pi^{t\frac{1}{2}} \\ 
\mat J^\zeta_{{\vct \epsilon}_{\mat R},k} &=\mat R_\pi[\vct\mu_k]_\times\mat\Sigma_\pi^{\mat R\frac{1}{2}}.
\end{alignat*}
%
Finally, we can compute
\[
\begin{alignedat}{2}
&
\begin{alignedat}{5}
    \mat\Sigma_k^\zeta\coloneqq& \mat J^\zeta_{\vct x,k}\mat\Sigma_k\mat J^{\zeta\top}_{\vct x,k}&+&\mat J^\zeta_{\vct \epsilon,k}\mat J^{\zeta\top}_{\vct \epsilon,k}\\
    =&\mat R_\pi\mat\Sigma_k\mat R_\pi^\top&+&\mat J^\zeta_{\vct \epsilon_{\mat R},k}\mat J^{\zeta\top}_{\vct \epsilon_{\mat R},k}&+&\mat J^\zeta_{\vct \epsilon_t,k}\mat J^{\zeta\top}_{\vct \epsilon_t,k}\\
    =&\mat R_\pi\mat\Sigma_k\mat R_\pi^\top&+&\mat R_\pi[\vct\mu_k]_\times\mat\Sigma_k^{\mat R}[\vct\mu_k]_\times^\top\mat R_\pi^\top&+&\mat R_\pi\mat\Sigma_\pi^t\mat R_\pi^\top
\end{alignedat}\\
    &\phantom{\mat\Sigma_k^\zeta{:}}=\mat R_\pi\left(\mat\Sigma_k+[\vct\mu_k]_\times\mat\Sigma_k^{\mat R}[\vct\mu_k]_\times^\top+\mat\Sigma_\pi^t\right)\mat R_\pi^\top\,.
\end{alignedat}
\]

\newpage

\section{Additional visualizations}

\begin{figure}[htp!]
    \centering
    \begin{subfigure}[t]{0.24\textwidth}
        \includegraphics[width=\textwidth]{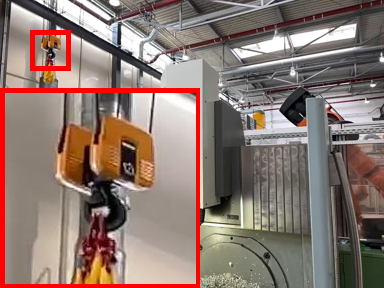}
    \end{subfigure}
    \begin{subfigure}[t]{0.24\textwidth}
        \includegraphics[width=\textwidth]{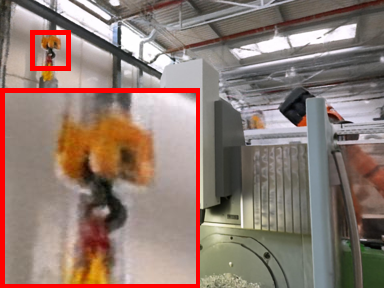}
    \end{subfigure}
    \begin{subfigure}[t]{0.24\textwidth}
        \includegraphics[width=\textwidth]{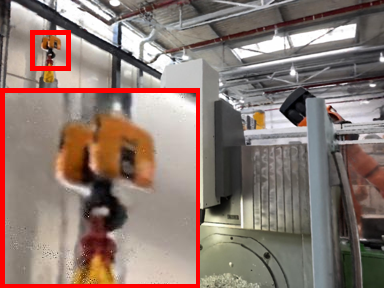}
    \end{subfigure}
    \begin{subfigure}[t]{0.24\textwidth}
        \includegraphics[width=\textwidth]{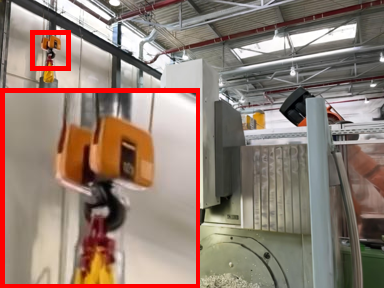}
    \end{subfigure}

    \begin{subfigure}[t]{0.24\textwidth}
        \includegraphics[width=\textwidth]{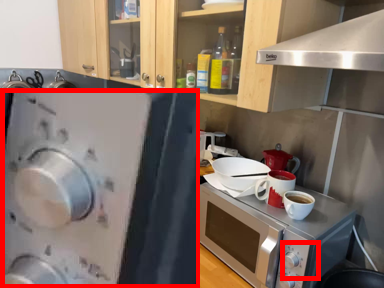}
    \end{subfigure}
    \begin{subfigure}[t]{0.24\textwidth}
        \includegraphics[width=\textwidth]{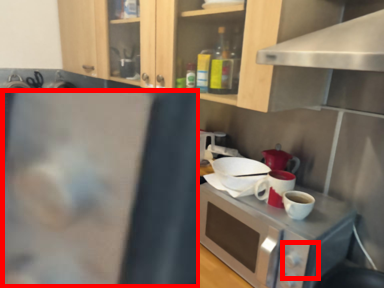}
    \end{subfigure}
    \begin{subfigure}[t]{0.24\textwidth}
        \includegraphics[width=\textwidth]{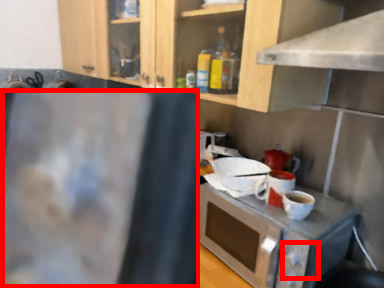}
    \end{subfigure}
    \begin{subfigure}[t]{0.24\textwidth}
        \includegraphics[width=\textwidth]{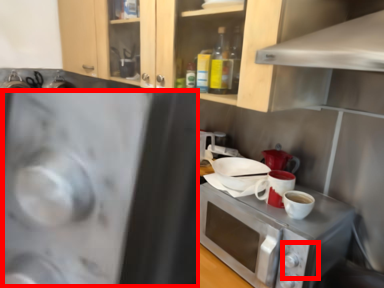}
    \end{subfigure}

    \begin{subfigure}[t]{0.24\textwidth}
        \includegraphics[width=\textwidth]{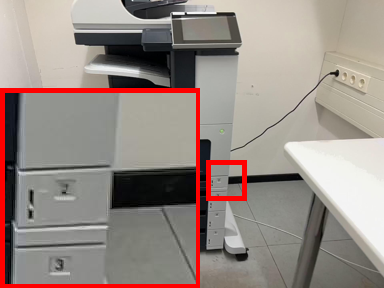}
    \end{subfigure}
    \begin{subfigure}[t]{0.24\textwidth}
        \includegraphics[width=\textwidth]{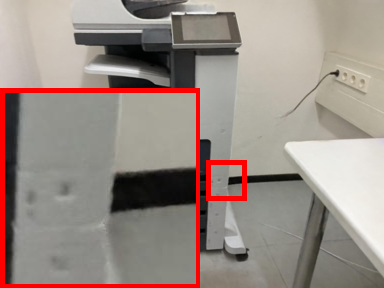}
    \end{subfigure}
    \begin{subfigure}[t]{0.24\textwidth}
        \includegraphics[width=\textwidth]{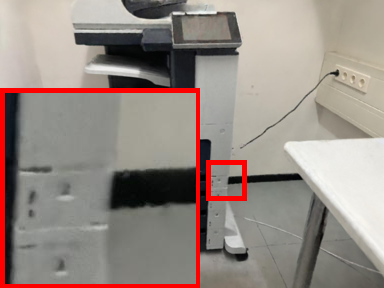}
    \end{subfigure}
    \begin{subfigure}[t]{0.24\textwidth}
        \includegraphics[width=\textwidth]{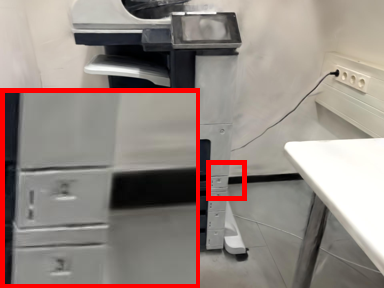}
    \end{subfigure} 

    \begin{subfigure}[t]{0.24\textwidth}
        \includegraphics[width=\textwidth]{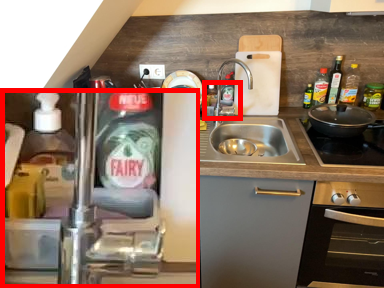}
    \end{subfigure}
    \begin{subfigure}[t]{0.24\textwidth}
        \includegraphics[width=\textwidth]{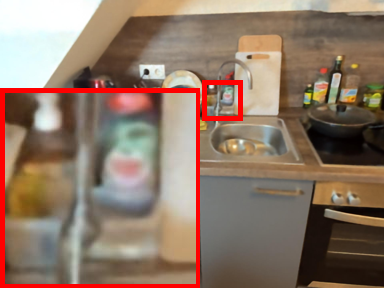}
    \end{subfigure}
    \begin{subfigure}[t]{0.24\textwidth}
        \includegraphics[width=\textwidth]{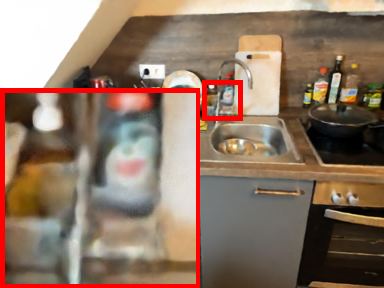}
    \end{subfigure}
    \begin{subfigure}[t]{0.24\textwidth}
        \includegraphics[width=\textwidth]{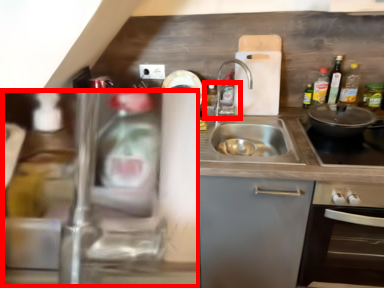}
    \end{subfigure}

    \begin{subfigure}[t]{0.24\textwidth}
        \includegraphics[width=\textwidth]{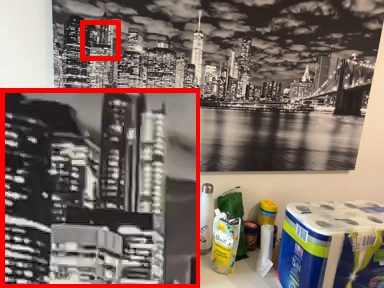}
    \end{subfigure}
    \begin{subfigure}[t]{0.24\textwidth}
        \includegraphics[width=\textwidth]{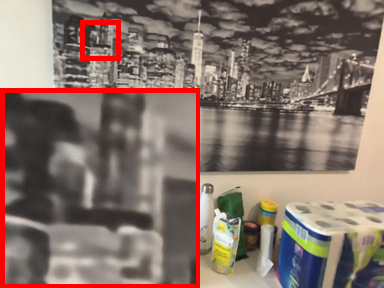}
    \end{subfigure}
    \begin{subfigure}[t]{0.24\textwidth}
        \includegraphics[width=\textwidth]{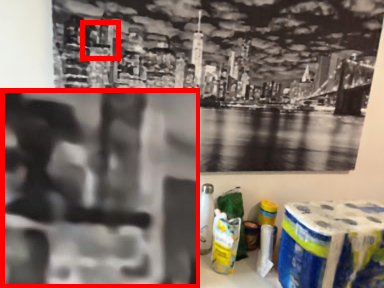}
    \end{subfigure}
    \begin{subfigure}[t]{0.24\textwidth}
        \includegraphics[width=\textwidth]{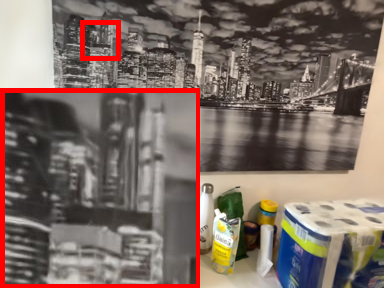}
    \end{subfigure}

    \begin{subfigure}[t]{0.24\textwidth}
        \includegraphics[width=\textwidth]{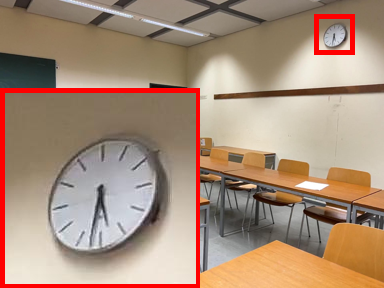}
        \caption{Ground truth}
    \end{subfigure}
    \begin{subfigure}[t]{0.24\textwidth}
        \includegraphics[width=\textwidth]{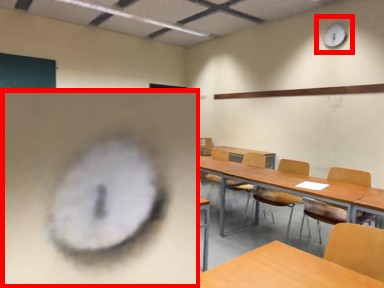}
        \caption{Mipnerf360~\cite{mipnerf360}}
    \end{subfigure}
    \begin{subfigure}[t]{0.24\textwidth}
        \includegraphics[width=\textwidth]{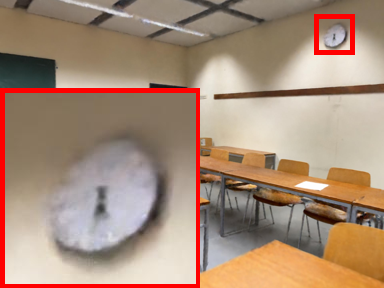}
        \caption{Nerfacto~\cite{nerfstudio}}
    \end{subfigure}
    \begin{subfigure}[t]{0.24\textwidth}
        \includegraphics[width=\textwidth]{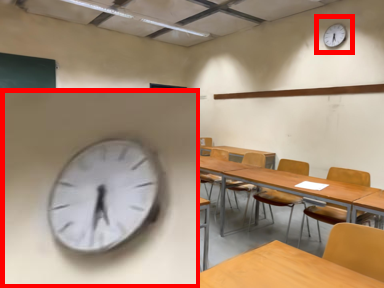}
        \caption{Ours}
    \end{subfigure}
    \caption{Comparison of our approach with state-of-the-art Mipnerf360~\cite{mipnerf360} and Nerfacto~\cite{nerfstudio}. All the methods are trained with per-image affine color transformation and pose optimization followed by test-time optimization of the per-image parameters on the test views.}
    \label{fig:quali}
\end{figure}

\begin{figure}[htp!]
    \centering
    \begin{subfigure}[t]{0.328\textwidth}
        \includegraphics[width=\textwidth]{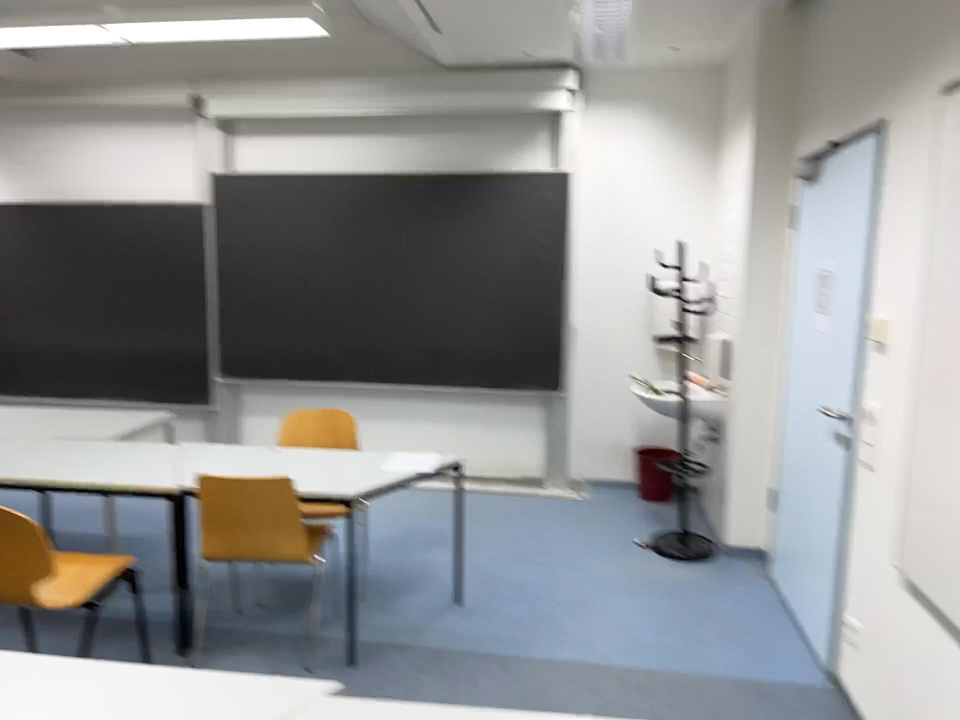}
    \end{subfigure}
    \begin{subfigure}[t]{0.328\textwidth}
        \includegraphics[width=\textwidth]{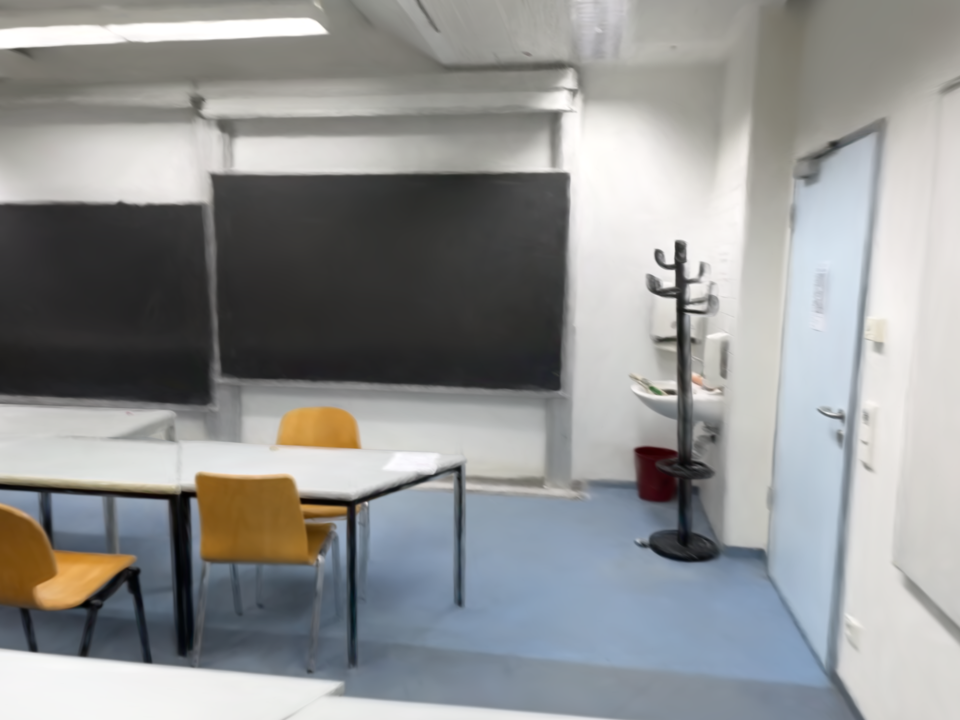}
    \end{subfigure}
    \begin{subfigure}[t]{0.328\textwidth}
        \includegraphics[width=\textwidth]{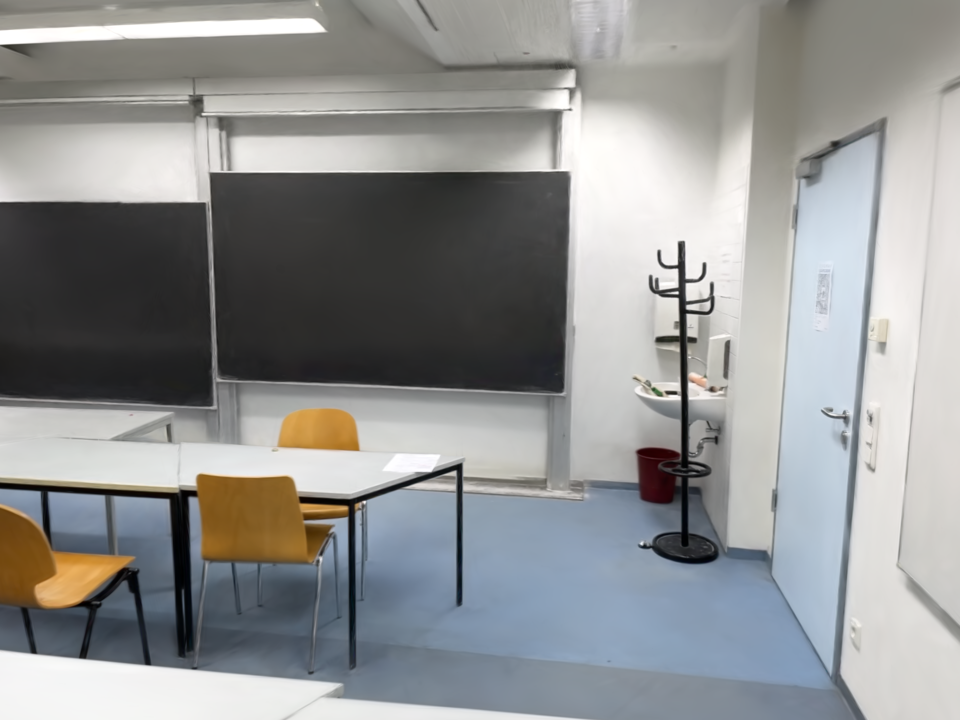}
    \end{subfigure}

    \begin{subfigure}[t]{0.328\textwidth}
        \includegraphics[width=\textwidth]{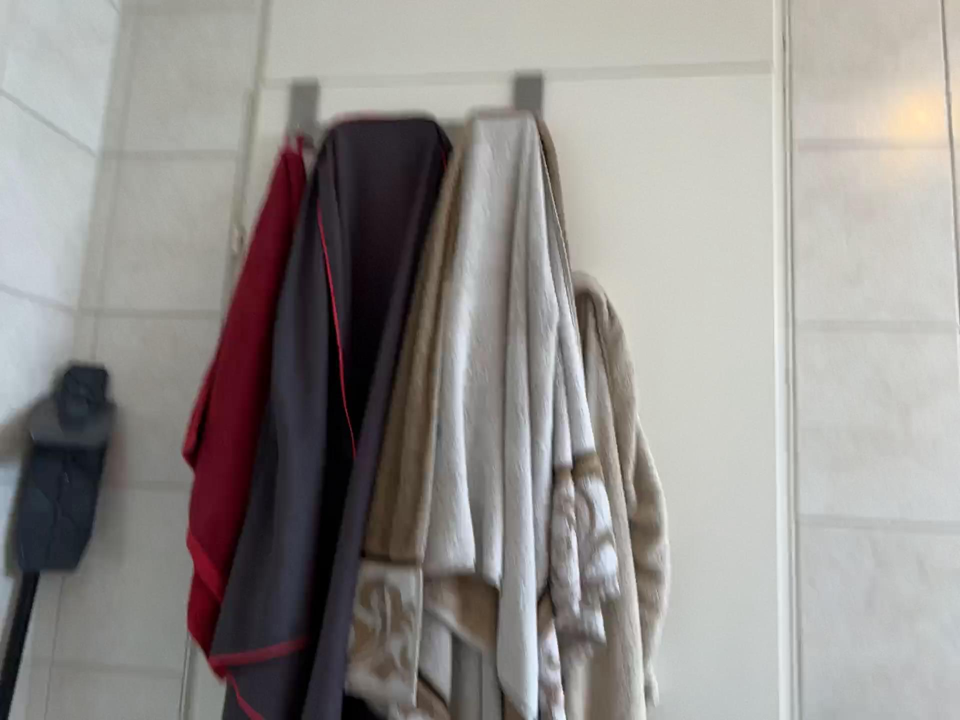}
    \end{subfigure}
    \begin{subfigure}[t]{0.328\textwidth}
        \includegraphics[width=\textwidth]{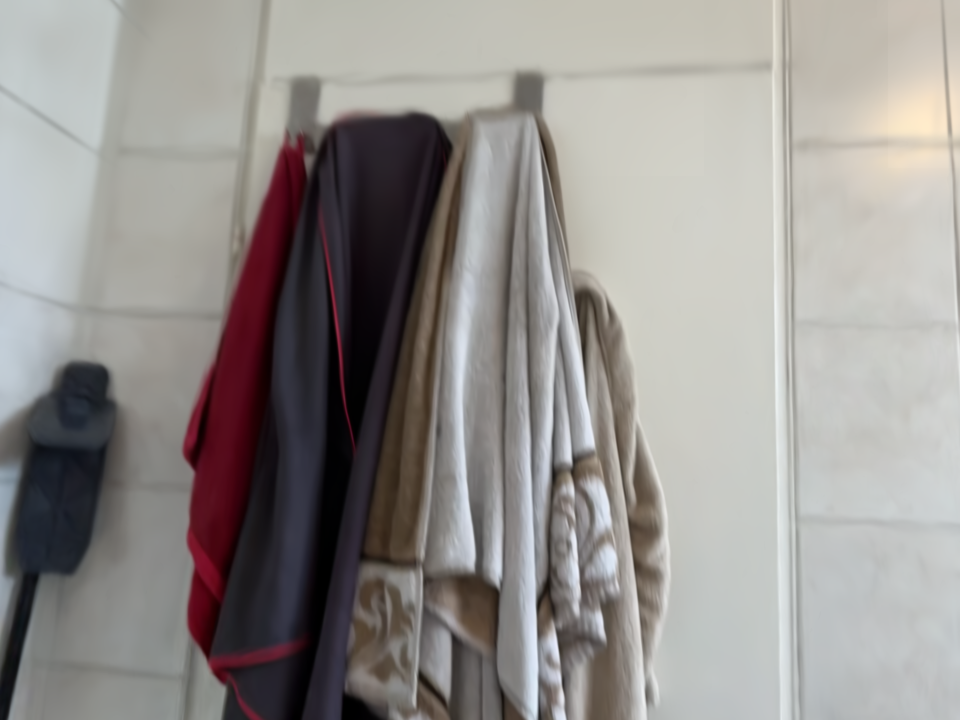}
    \end{subfigure}
    \begin{subfigure}[t]{0.328\textwidth}
        \includegraphics[width=\textwidth]{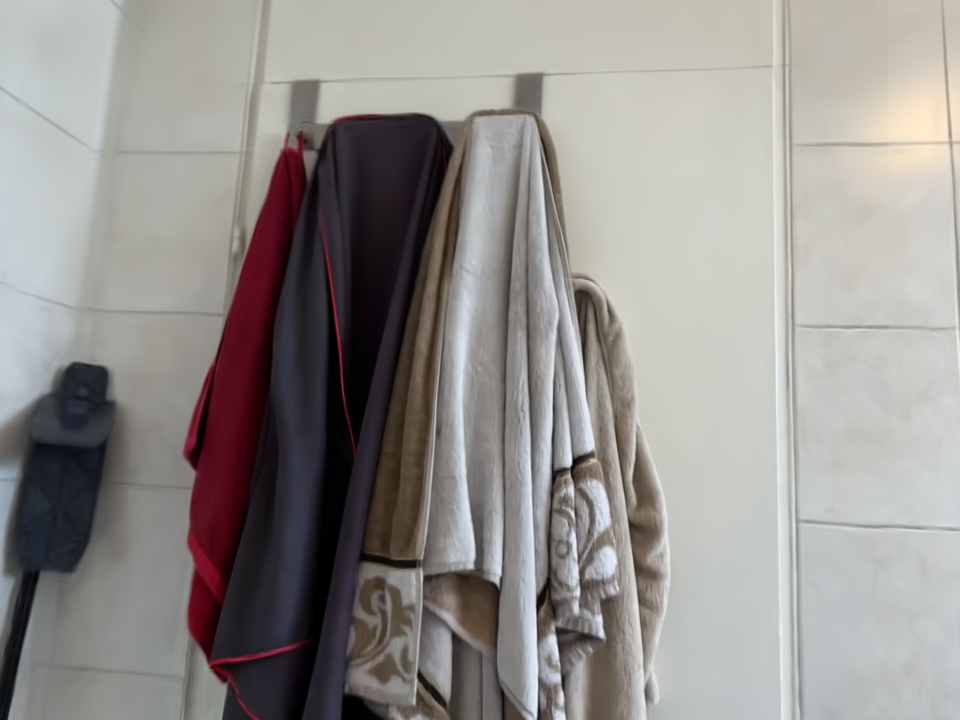}
    \end{subfigure}

    \begin{subfigure}[t]{0.328\textwidth}
        \includegraphics[width=\textwidth]{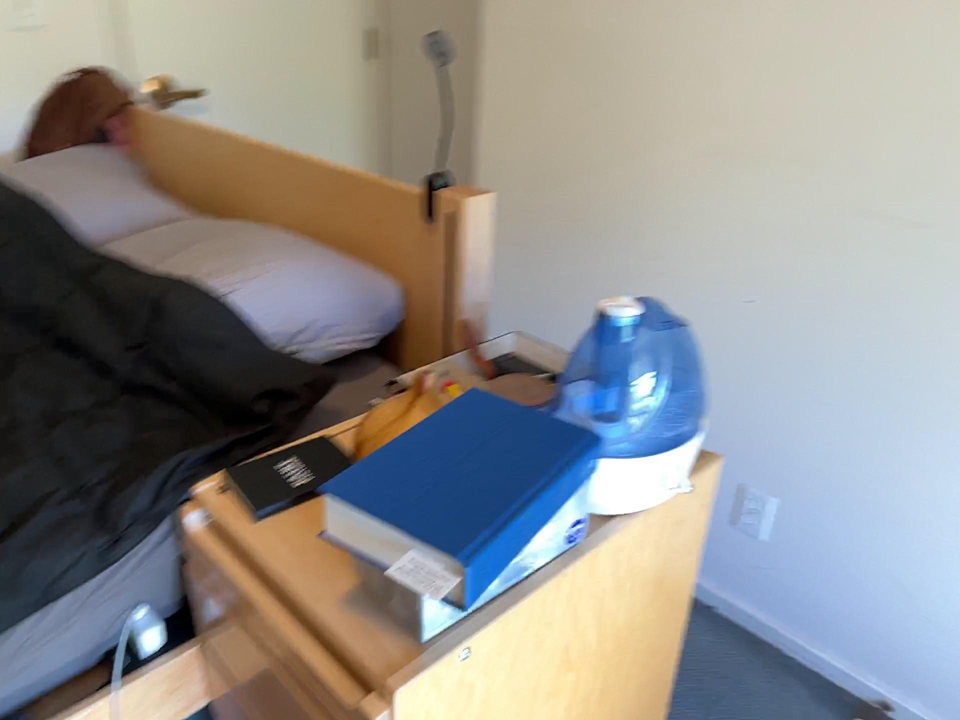}
    \end{subfigure}
    \begin{subfigure}[t]{0.328\textwidth}
        \includegraphics[width=\textwidth]{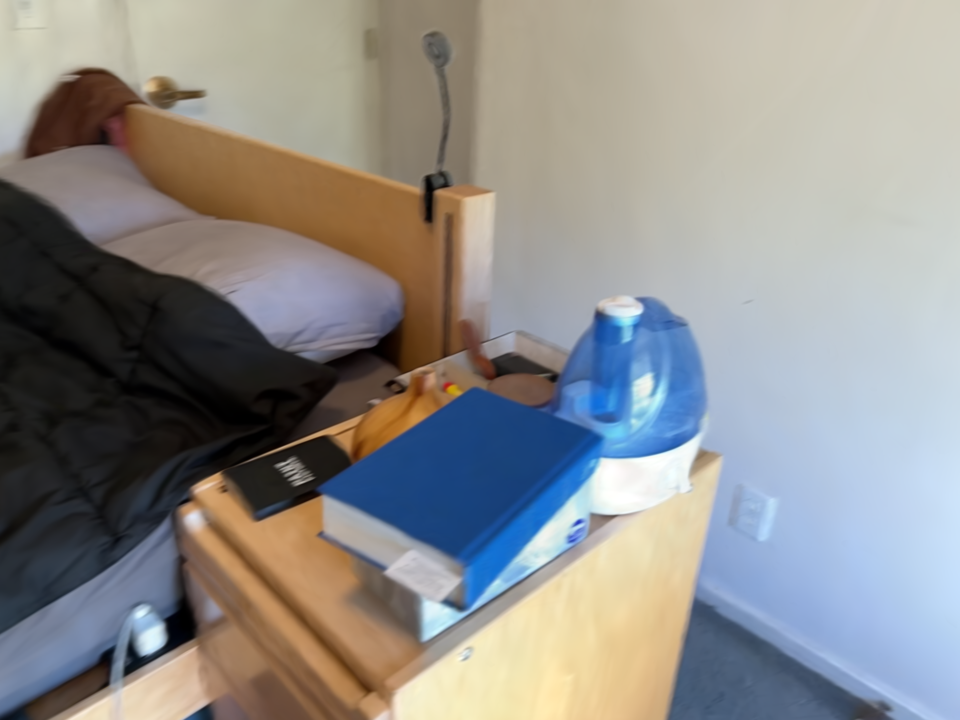}
    \end{subfigure}
    \begin{subfigure}[t]{0.328\textwidth}
        \includegraphics[width=\textwidth]{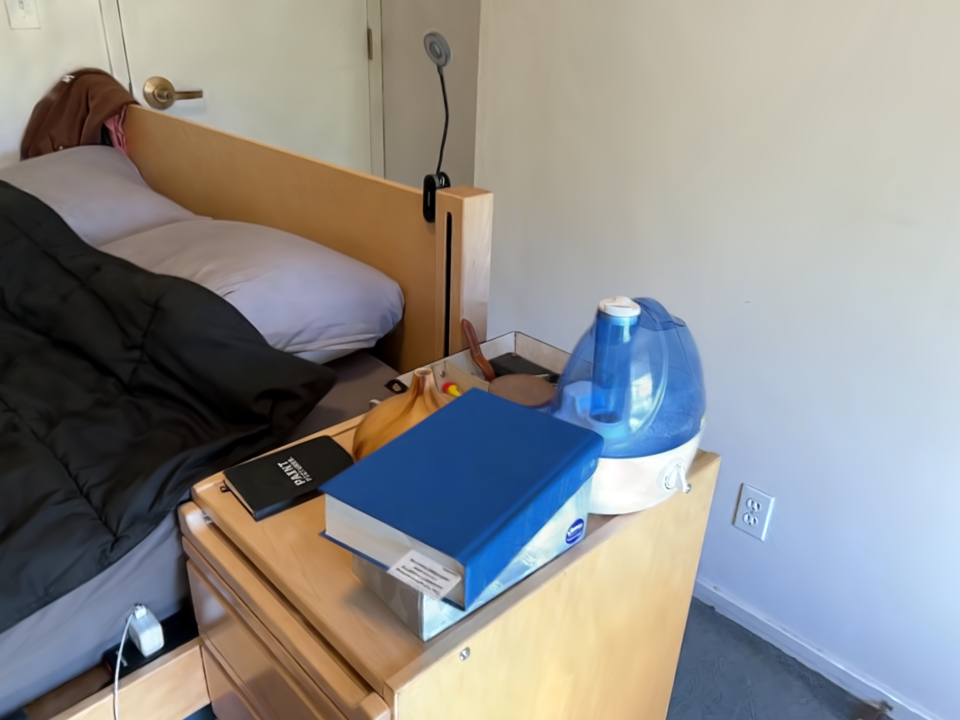}
    \end{subfigure}

    \begin{subfigure}[t]{0.328\textwidth}
        \includegraphics[width=\textwidth]{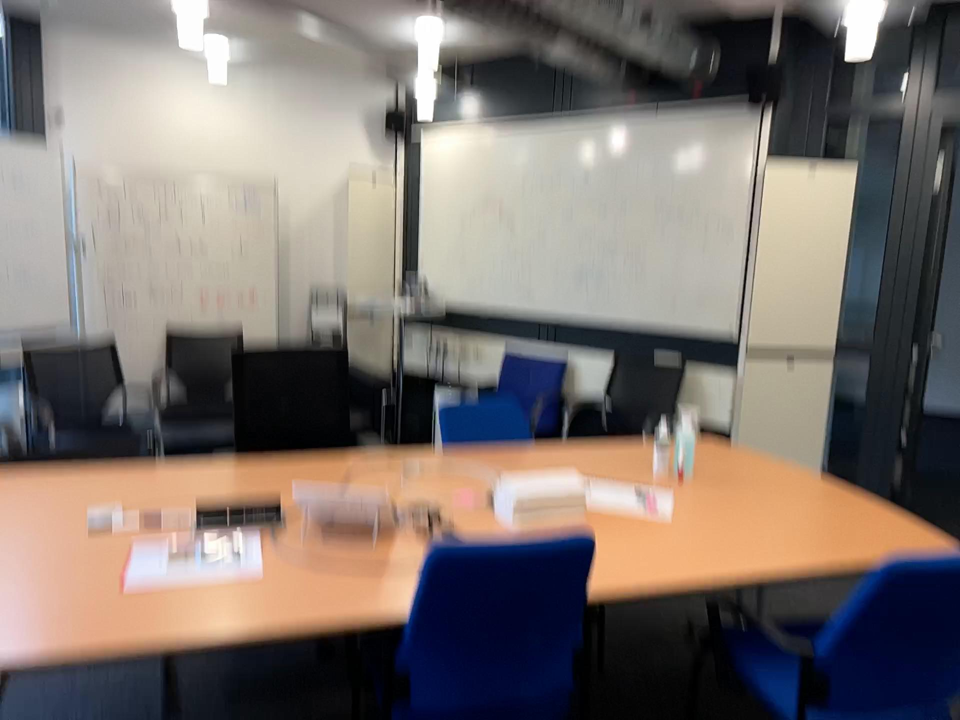}
    \end{subfigure}
    \begin{subfigure}[t]{0.328\textwidth}
        \includegraphics[width=\textwidth]{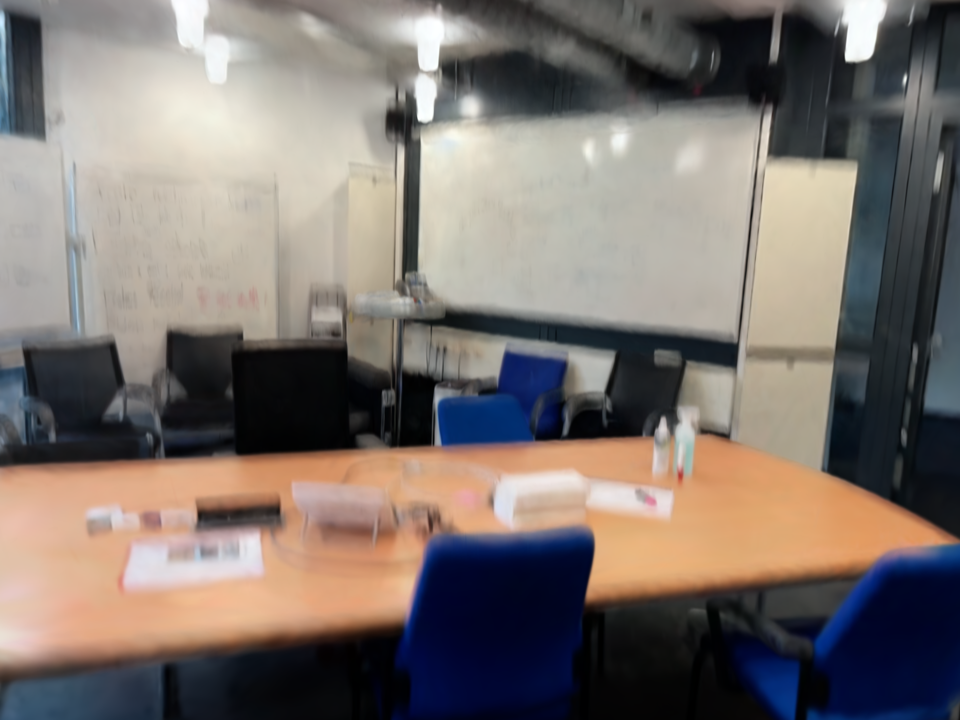}
    \end{subfigure}
    \begin{subfigure}[t]{0.328\textwidth}
        \includegraphics[width=\textwidth]{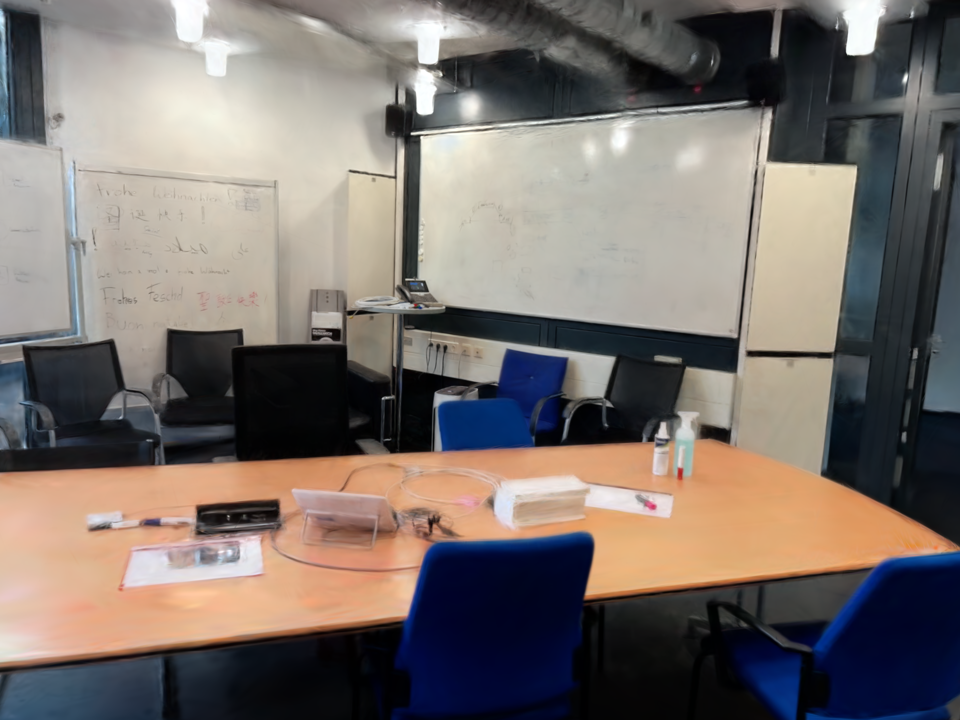}
    \end{subfigure}

    \begin{subfigure}[t]{0.328\textwidth}
        \includegraphics[width=\textwidth]{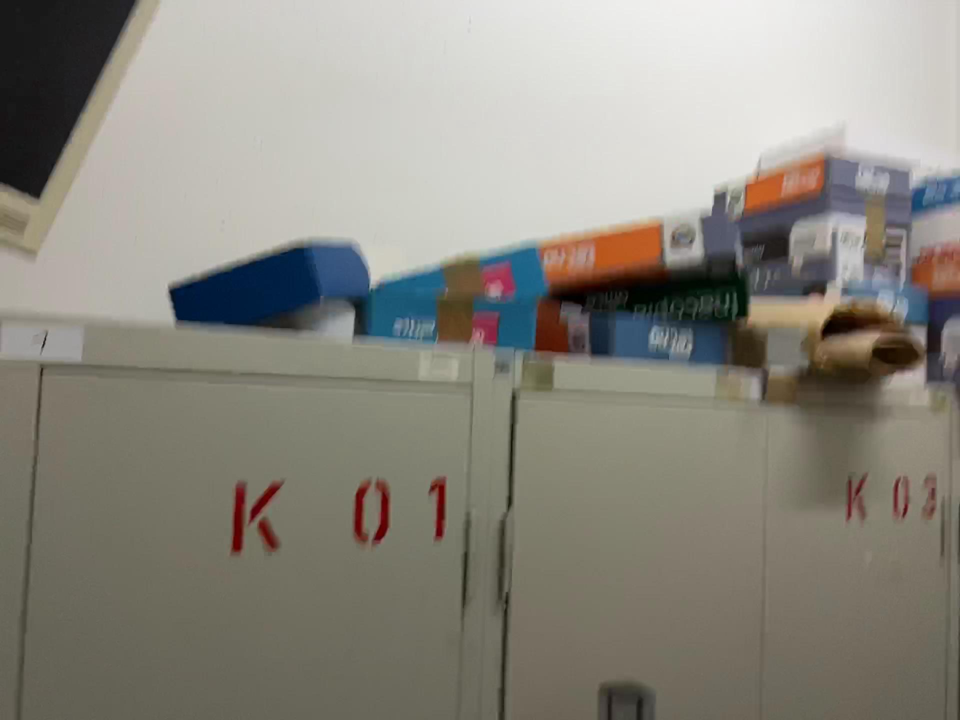}
        \caption{Blurry training view}
    \end{subfigure}
    \begin{subfigure}[t]{0.328\textwidth}
        \includegraphics[width=\textwidth]{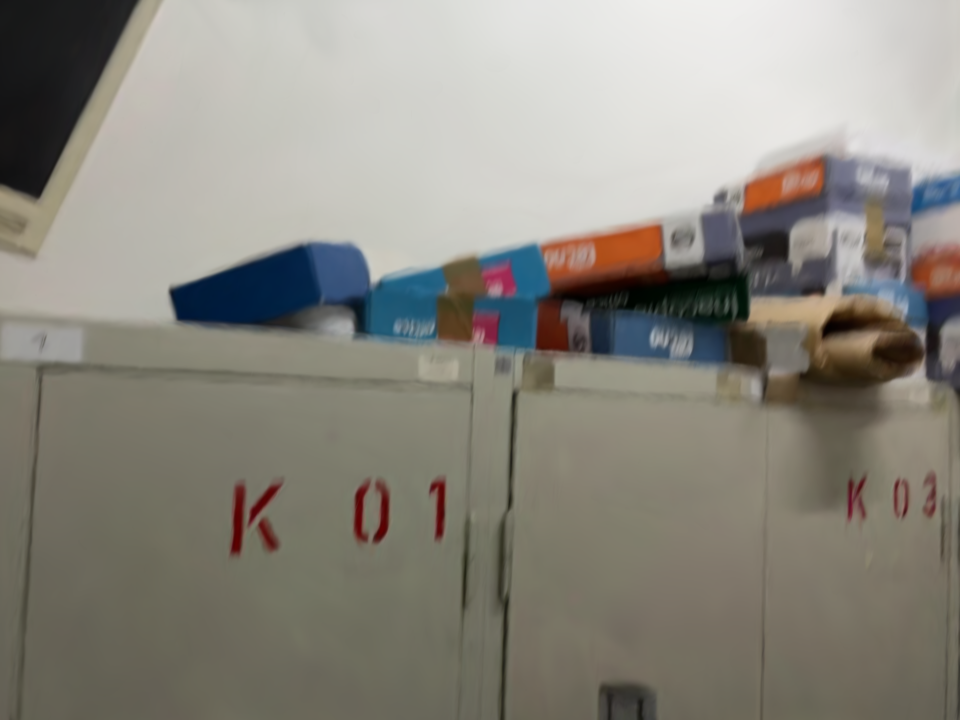}
        \caption{Ours with blurring}
    \end{subfigure}
    \begin{subfigure}[t]{0.328\textwidth}
        \includegraphics[width=\textwidth]{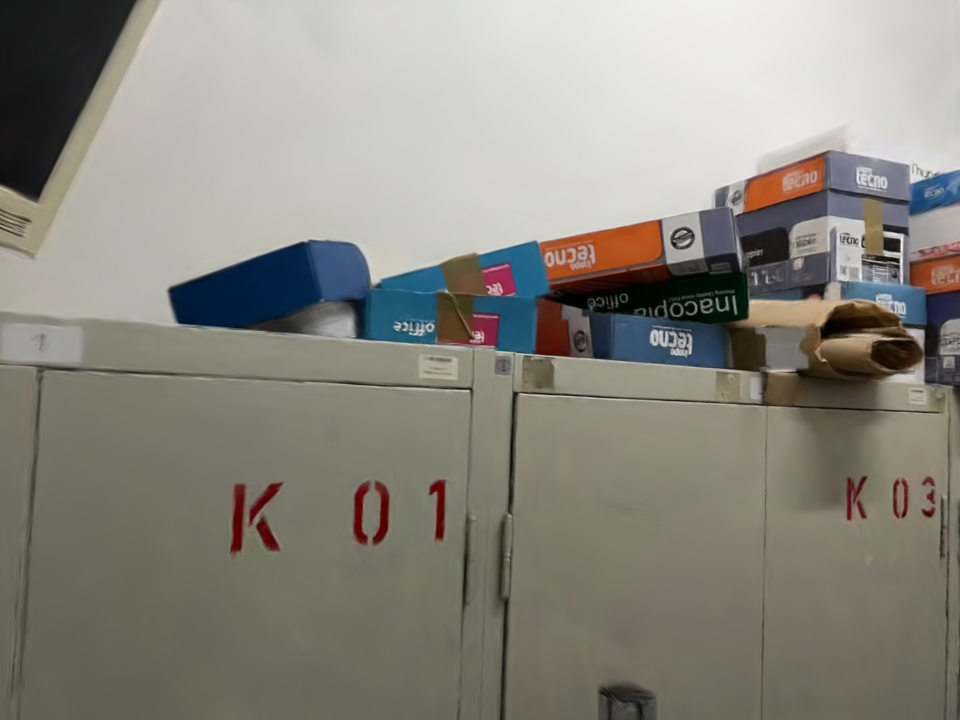}
        \caption{Ours without blurring}
    \end{subfigure}
    \caption{Side by side comparison of a blurry training view from Scannet++ (a), rendering with blurring enabled (b) and rendering from the same model but with blurring disabled (c). Our model is able to copy a blurry training view while keeping a sharp representation.}
    \label{fig:visu_blur}
\end{figure}

\end{document}